%% file: main.tex
\documentclass[10pt]{article} 

\usepackage[preprint]{rlj} 

\usepackage{amssymb}            
\usepackage{mathtools}          
\usepackage{mathrsfs}           
\usepackage{graphicx}           
\usepackage{subcaption}         
\usepackage[space]{grffile}     
\usepackage{url}                
\usepackage{lipsum}             
\usepackage{amsthm}
\usepackage{cleveref}

\newtheorem{theorem}{Theorem}
\newtheorem{corollary}[theorem]{Corollary}
\newtheorem{proposition}[theorem]{Proposition}

\usepackage{dsfont}
\usepackage{booktabs}
\usepackage{makecell}

\usepackage{color}

\usepackage{ifthen}
\newboolean{DisplayComments}
\setboolean{DisplayComments}{false}
\DeclareRobustCommand{\aj}[1]{\ifthenelse{\boolean{DisplayComments}}{{\color{red} (AJ: #1)}}{}}
\DeclareRobustCommand{\raj}[1]{\ifthenelse{\boolean{DisplayComments}}{{\color{blue} (Raj: #1)}}{}}
\DeclareRobustCommand{\yc}[1]{\ifthenelse{\boolean{DisplayComments}}{{\color{violet} (YC: #1)}}{}}

\title{Safe RLHF Beyond Expectation: Stochastic Dominance for Universal Spectral Risk Control}

\setrunningtitle{Safe RLHF Beyond Expectation}

\author{Yaswanth Chittepu\textsuperscript{1,$\dagger$}, Ativ Joshi\textsuperscript{1}, Rajarshi Bhattacharjee\textsuperscript{1}, Scott Niekum\textsuperscript{1}}

\emails{\{ychittepu, atjoshi, rbhattacharj, sniekum\}@umass.edu}

\affiliations{
  $^{1}$\textbf{University of Massachusetts Amherst}\\
}

\contribution{
  We introduce Risk-sensitive Alignment via Dominance (RAD), a Safe RLHF objective that constrains the first-order stochastic dominance (FSD) of the learned policy’s cost distribution relative to a reference policy, rather than constraining only expected cost.
}
{
  Prior Safe RLHF \citep{dai2023safe} and HC-RLHF \citep{chittepu2025reinforcement} formulations impose scalar expected-cost or high-confidence expected-cost constraints, while we position RAD as bringing stochastic-dominance control into the safe-RLHF framework for policy-induced cost distributions.
}

\contribution{
  We give a practical optimization procedure for this dominance objective by using an asymmetric quantile-gap surrogate, a nonparametric empirical quantile-particle representation of policy-induced cost distributions, and an entropically regularized optimal-transport formulation that yields a differentiable REINFORCE-style policy-gradient estimator.
}
{
  While the OT view of the FSD surrogate and Sinkhorn-based regularization have been explored \citep{melnyk2024distributional}, we adapt them to the sample-based Safe RLHF setting together with the resulting gradient estimator for optimizing the dominance-constrained objective.
}

\contribution{
  We introduce quantile-weighted FSD constraints and show that they provide universal control over a broad class of Spectral Risk Measures (SRMs), enabling tunable risk sensitivity.
}
{
SRMs \citep{acerbi2002spectral} form a broad class of coherent \citep{artzner1999coherent}, law-invariant risk functionals that aggregate quantile costs using a weighted spectrum over
confidence levels.
}

\contribution{
  Empirically, we show that RAD improves the harmlessness of model responses over baselines (Safe-RLHF, SFT) and generalizes better to out-of-distribution harmlessness evaluations, while maintaining competitive helpfulness relative to Safe-RLHF \citep{dai2023safe}.
}
{
We use the BeaverTails dataset \citep{ji2023beavertailsimprovedsafetyalignment} 
following \citet{dai2023safe}, and evaluate the helpfulness and harmlessness of RAD, Safe-RLHF, and SFT models, on its test split. We additionally evaluate out-of-distribution harmlessness on HarmBench 
\citep{mazeika2024harmbenchstandardizedevaluationframework}.
}

\keywords{Reinforcement Learning from Human Feedback (RLHF), Safety, Alignment, Stochastic Dominance, Optimal Transport, Spectral Risk Measures} 

\summary{

Safe RLHF typically enforces safety through expected cost constraints, but the expectation captures only a single statistic of the cost distribution and fails to account for distributional uncertainty, particularly under heavy tails or rare catastrophic events. We proposes Risk-sensitive Alignment via Dominance (RAD), a safe alignment framework that replaces expected-cost constraints with first-order stochastic dominance (FSD) constraints on the full cost distribution relative to a reference policy, implemented through an entropically regularized optimal transport objective. Furthermore, we introduce quantile-weighted dominance constraints and show that they provide universal control over a broad class of Spectral Risk Measures, enabling tunable risk sensitivity.

}

\begin{document}

\makeCover  
\maketitle  

\begin{abstract}

  Safe Reinforcement Learning from Human Feedback (RLHF) typically enforces safety through expected cost constraints, but the expectation captures only a single statistic of the cost distribution and fails to account for distributional uncertainty, particularly under heavy tails or rare catastrophic events. This limitation is problematic when robustness and risk sensitivity are critical. Stochastic dominance offers a principled alternative by comparing entire cost distributions rather than just their averages, enabling direct control over tail risks and potential out-of-distribution failures that expectation-based constraints may overlook. In this work, we propose Risk-sensitive Alignment via Dominance (RAD), a novel alignment framework that replaces scalar expected cost constraints with First-Order Stochastic Dominance (FSD) constraints. We operationalize this constraint by comparing the target policy's cost distribution to that of a reference policy within an Optimal Transport (OT) framework, using entropic regularization and Sinkhorn iterations to obtain a differentiable and computationally efficient objective for stable end-to-end optimization. Furthermore, we introduce quantile-weighted FSD constraints and show that weighted FSD universally controls a broad class of Spectral Risk Measures (SRMs), so that improvements under weighted dominance imply guaranteed improvements in the corresponding spectral risk. This provides a principled mechanism for tuning a model's risk profile via the quantile weighting function. Empirical results demonstrate that RAD improves harmlessness over baselines while remaining competitive in helpfulness, and exhibits greater robustness on out-of-distribution harmlessness evaluations.

\end{abstract}

\section{Introduction}
\label{sec: introduction}
\input{sections/introduction}

\section{Background and Problem Setting}
\label{sec: background}
\input{sections/background}

\section{Risk-sensitive Alignment via Dominance}
\label{sec: method}
\input{sections/method}

\section{Empirical Results}
\label{sec: results}
\input{sections/experiments}

\section{Related Work}
\label{sec: relatedwork}

\input{sections/relatedwork}

\section{Conclusion}
\label{sec: conclusion}
\input{sections/conclusion}

\section*{Acknowledgements}

This work has taken place in the Safe, Correct, and Aligned Learning and Robotics Lab (SCALAR) at The University of Massachusetts Amherst. SCALAR research is supported in part by the NSF (IIS-2437426) and Open Philanthropy. The computational resources for this work were provided by the University of Massachusetts Amherst's partnership with the Unity Research Computing Platform, a multi-institutional cluster led by the University of Massachusetts and the University of Rhode Island.

Scott Niekum holds concurrent appointments as an Associate Professor at the University of Massachusetts Amherst and as an Amazon Scholar. This paper describes work performed at the University of Massachusetts Amherst and is not associated with Amazon.
\bibliography{main}
\bibliographystyle{rlj}

\beginSupplementaryMaterials
\appendix
\input{sections/supplementary}

\end{document}

%% file: sections/introduction.tex

Large Language Models (LLMs) are now used across a wide range of applications, so it is important that their outputs are helpful, reliable, and safe. This need is especially acute in high-stakes domains such as legal reasoning (\cite{katz2024gpt}), medical consultation (\cite{yang2022large,moor2023foundation}), and educational support (\cite{kasneci2023chatgpt,kung2023performance}), where harmful generations—including toxicity and misinformation—can have serious consequences (\cite{gehman2020realtoxicityprompts,weidinger2021ethical,ganguli2022red}).

Broadly, we want LLMs to be \textit{helpful} and aligned with human preferences, while being \textit{harmless}, i.e. prevent toxic outputs. But there may be scenarios when users may request assistance with potentially harmful activities (\cite{Bai2022ConstitutionalAH,glaese2022improving}). Thus, \emph{harmlessness} and \emph{helpfulness} can conflict. The standard technique for aligning LLMs with human preferences, Reinforcement Learning from Human Feedback (RLHF), typically optimizes a single reward model for both objectives (\cite{Ouyang2022TrainingLM,bai2022traininghelpfulharmlessassistant}), or heuristically mixes separate reward models (\cite{glaese2022improving,touvron2023llama,mu2024rule}). This creates trade-offs: stronger harmlessness can lead to refusals, while stronger helpfulness can increase unsafe outputs (\cite{bai2022traininghelpfulharmlessassistant}). Recent work addresses this by decoupling preference data and enforcing harmlessness as a constraint, an approach known as Safe RLHF (\cite{dai2023safe}). However, Safe RLHF typically typically constrains the expected cost of a policy, providing no guarantees on worst-case or tail outcomes.  This is inadequate in high-stakes deployments where tail risk is critical — such as code generation \cite{pearce2021asleepkeyboardassessingsecurity} and open-ended dialogue where rare but severe harms including toxic generations and private data leakage can impact users
\citep{perez2022redteaminglanguagemodels, gehman2020realtoxicityprompts}.




\yc{Also, I changes the flow of the old paragraph, and incorporated scott's comments. Lmk if this seems good or not}\raj{we already talk about expected cost contraints above. Moved the first line to the pargarph above since it is more expressive.}
Thus, we argue that a stronger notion of constraining the cost of the policy is needed: the cost distribution of the learned policy should be \emph{stochastically smaller} than that of the reference policy, not merely cheaper on average. \raj{added a linbe which explains what it means to be stcoshstically smaller} This means that the learned policy should assign less probability to high-cost outcomes across the distribution, not just reduce the mean cost. This motivates \textbf{Risk-sensitive Alignment via Dominance (RAD)}, which enforces a First-Order Stochastic Dominance (FSD) constraint on the cost distribution of the learned policy relative to a reference policy.

A key insight of RAD is that by reweighting quantiles of the FSD objective, we recover Spectral Risk Measures (SRMs) — a family of risk measures defined as a weighted integral of the cost quantiles,
$\int w(q) Q_X(q) dq$, where $Q_X(q)$ is the quantile function of the cost distribution and the weighting function $w(q)$ allows
practitioners to express diverse risk preferences over the cost distribution. For instance, concentrating weight on upper quantiles recovers tail-sensitive measures such as CVaR, while uniform weighting recovers the mean — both expressible within our unified framework. This is practically valuable: a medical deployment may demand near-zero tolerance for harmful outputs, while a general assistant
may tolerate a more permissive tradeoff, and both are accommodated by choosing $w$ appropriately.

Optimizing FSD constraints directly is challenging, so we relax them to an asymmetric FSD-violation surrogate that aggregates positive quantile gaps (see \Cref{eq:fsd_loss}). We interpret stochastic
dominance through Optimal Transport (OT) and derive an efficient REINFORCE-style policy gradient using Sinkhorn iterations, yielding an end-to-end differentiable approach. Empirically, RAD yields models that are more robust to safety violations while achieving competitive helpfulness across multiple spectral risk measures relative to Safe RLHF.
\yc{IS this ok, given the results?}\raj{modified a bit}



\aj{verify}
Our main contributions are: (i) we formulate safe alignment as a dominance-constrained objective over the full cost distribution rather than only its expectation, (ii) we derive a RAD policy-gradient estimator using entropic OT and Sinkhorn iterations, making optimization of the objective end-to-end differentiable, (iii) we connect quantile reweighting in our framework to SRMs, enabling controllable risk-sensitive alignment profiles, and (iv) through extensive experiments, we show that RAD improves robustness to safety violations while maintaining competitive helpfulness relative to expected cost based baselines. \aj{extra} The rest of the paper is organized as follows: \Cref{sec: background} reviews Safe RLHF, stochastic dominance, and SRMs; \Cref{sec: method} presents RAD and its optimization procedure; \Cref{sec: results} reports empirical evaluations safety--helpfulness trade-offs; and \Cref{sec: relatedwork,sec: conclusion} is related work and conclusion respectively.
\aj{Can write one or two lines highlighting the conclusion of experiments.}\raj{changed second point. Replacing unregularized with regaularized OT is not really an original contribution and pretty standard. But we show how to estimate the gradients which could be a new contribution} \yc{Add experimental conclusions as a point in our contribution} \raj{added experimental contributional.}

%% file: sections/background.tex

\subsection{Reinforcement Learning from Human Feedback (RLHF)}

The most popular method of aligning LLMs with human preferences is Reinforcement Learning from Human Feedback (RLHF). The standard RLHF pipeline consists of three phases as described in \cite{Christiano2017DeepRL,Ouyang2022TrainingLM}. The first phase is a supervised fine-tuning (SFT) phase where the outputs of a pre-trained model are aligned with human responses. This is followed by a reward modeling phase, where a reward model is trained using human preferences and finally, a reinforcement learning phase where the model is optimized using the learned reward function. Formally, let  $x \sim D_x$  denote prompts and  $y \sim \pi_\theta(\cdot \mid x)$  denote model responses. SFT produces a reference policy  $\pi_{\mathrm{sft}}$, and we set  $\pi_{\mathrm{ref}} := \pi_{\mathrm{sft}}$  unless stated otherwise. A reward model  $r_\phi(x,y)$  is learned from preference comparisons using the Bradley-Terry model \citep{bradley1952rank}. Details are standard and deferred to \Cref{supp:RLHF}.

RLHF then optimizes a KL-regularized objective to prevent reward overoptimization and preserve language quality \citep{Ouyang2022TrainingLM,stiennon2022learningsummarizehumanfeedback,Gao2022ScalingLF,Rafailov2024ScalingLF}:
\begin{align}
  \max_\theta\;\mathbb{E}_{x \sim D_x,\, y \sim \pi_\theta(\cdot\mid x)}\!\big[r_\phi(x,y)\big]\;-\;\beta\,D_{\mathrm{KL}}\!\Big(\pi_\theta(\cdot\mid x)\,\Vert\,\pi_{\mathrm{ref}}(\cdot\mid x)\Big),
\end{align}
equivalently maximizing the KL-regularized reward  $\tilde r(x,y)=r_\phi(x,y)-\beta\log\frac{\pi_\theta(y\mid x)}{\pi_{\mathrm{ref}}(y\mid x)}$. In practice, this objective is optimized with PPO/GRPO/REINFORCE-style methods \citep{williams1992simple,ahmadian2024basicsrevisitingreinforcestyle,shao2024deepseekmathpushinglimitsmathematical,schulman2017proximalpolicyoptimizationalgorithms}.


\subsection{Safe RLHF}
Standard RLHF optimizes a single reward function learned from human preferences, which may be inadequate when attempting to balance competing goals such as \emph{helpfulness} and \emph{harmlessness}. In contrast, Safe RLHF \citep{dai2023safe} separates helpfulness and harmlessness by learning (i) a reward model  $r_\phi(x,y)$, (parameterized by $\phi$ and for input prompt $x$ and model output $y$)  from helpfulness preferences and (ii) a cost model  $c_\psi(x,y)$, (parameterized by $\psi$)  from harmfulness preferences, and then solving the expected-cost constrained RL problem:
\begin{align}
  \max_\theta\;\mathbb{E}\big[r_\phi(x,y)\big] - \beta D_{\mathrm{KL}}(\pi_\theta\Vert \pi_{\mathrm{ref}}) \quad\text{s.t.}\quad \mathbb{E}\big[c_\psi(x,y)\big]\le \tau.
  \label{eq:safe-rlhf-obj}
\end{align}

High-confidence variants like HC-RLHF \citep{chittepu2025reinforcement} replace the expectation constraint with a probabilistic guarantee using the Seldonian framework \citep{thomas2019preventing}.

\subsection{First-Order Stochastic Dominance}
\label{sec:fosd}

\aj{cite for FSD} \raj{incorporated Scott's suggestion about describing our approach more accuartely.}
The expected cost constraint in Safe RLFH often proves inadequate in settings where controlling the tail risk is crucial. A more robust method for controlling tail risk is ensuring that the learned policy's cost is stochastically smaller than the cost of the reference policy i.e. less probability must be assigned by the learned policy to high cost outcomes compared to the reference policy. We now describe how we do this formally. Let $F_{X}$ and $Q_{X}$ denote the Cumulative Distribution Function (CDF) and the Quantile Function (Inverse CDF) of a random variable $X$.
For real-valued random variables  $X,Y$, $X$ is said to have first-order stochastic dominance (FSD) on $Y$  (denoted by $X \succeq_{\mathrm{FSD}} Y$)  iff  $F_X(r)\le F_Y(r)$  for all  $r$ \citep{dai2023learning}. Equivalently, we also have $Q_X(q)\ge Q_Y(q)$  for all  $q\in[0,1]$.

Practically, a drawback of FSD is that it only imposes a partial ordering over distributions, so two random variables may be incomparable in terms of FSD. Hence, we relax the definition to an objective that measures the extent to which $X$ falls below $Y$ across quantiles:
\begin{equation}
  \mathcal{L}_{\mathrm{FSD}}(X,Y)
  := \int_{0}^{1} \big(Q_Y(q)-Q_X(q)\big)_+\, dq,
  \label{eq:fsd_loss}
\end{equation}
where $(x)_+=\max(x,0)$ denotes the ReLU function. Note that for two random variables $X$ and $Y$ with different distributions, an FSD objective of $0$ implies that X dominates Y, i.e.  $X \succeq_{\mathrm{FSD}} Y$. More generally,  $\mathcal{L}_{\mathrm{FSD}}(X,Y)$  aggregates the positive quantile gaps  $(Q_Y(q)-Q_X(q))_+$  and therefore quantifies the extent to which  $X$  fails to dominate  $Y$ or $Y$ dominates $X$; it does not by itself certify  $Y\succeq_{\mathrm{FSD}}X$.

In our case, we set $X$ to be the cost distribution $C_{\pi_{\theta}}$ under our learned policy $\Pi_{\theta}$ and $Y$ to be the cost distribution $C_{\pi_{\mathrm{ref}}}$ under the reference policy $\Pi_{\mathrm{ref}}$. Thus, $\mathcal{L}_{\mathrm{FSD}}(C_{\pi_{\theta}},C_{\pi_{\mathrm{ref}}})$ quantifies by how much the cost distribution of the reference policy dominates the cost distribution of the learned policy\footnote{Note that $\mathcal{L}_{\mathrm{FSD}}(X,Y)$ is not a distance because it is not symmetric and can be zero even when $X$ and $Y$ have different distributions. Also $\mathcal{L}_\mathrm{FSD}(X,Y) + \mathcal{L}_\mathrm{FSD}(Y,X) = \mathcal{W}_{1}(X,Y)$ where $\mathcal{W}_{1}(X,Y)$ is the 1-Wassertsein distance }.

\subsection{Optimal Transport}\label{subsec:ot}


Optimal transport \citep{peyre2020computationaloptimaltransport} studies the problem of optimally transforming a distribution $\mu$ into another distribution $\nu$ under a cost function
$c(x,y) : \mathcal{X} \times \mathcal{Y} \to \mathbb{R}_+$, which represents the cost of moving a unit mass from $x$ to $y$.

Let $\mu = \sum_{i=1}^N a_i\,\delta_{x_i}$ and $\nu = \sum_{j=1}^M b_j\,\delta_{y_j}$ be two empirical (atomic) distributions on $\mathbb{R}^{d}$, where $x_i, y_j \in \mathbb{R}^d$ are support points (“particles”), $a \in \Delta^N$ and $b \in \Delta^M$ are nonnegative weights summing to 1 ($\Delta^N$ denotes the $N$-dimensional probability simplex), and $\delta_z$ is a Dirac mass at $z$.

A \emph{transport plan} (a.k.a. coupling) from $\mu$ to $\nu$ is a nonnegative matrix $P \in \mathbb{R}_+^{N \times M}$ whose row and column sums match the marginals. Intuitively, $P_{ij}$ represents the amount of probability mass moved from $x_i$ to $y_j$. The set of admissible couplings can be represented as:
\begin{equation}
  \Pi(\mu,\nu) := \Bigl\{ P \in \mathbb{R}_+^{N \times M} \;:\; P \mathbf{1}_M = a, \;\; P^\top \mathbf{1}_N = b \Bigr\}.
  \label{eq:transport-plans}
\end{equation}

Given a ground cost function $c:\mathbb{R}^d \times \mathbb{R}^d \to \mathbb{R}_+$, define the cost matrix $C_{ij} = c(x_i, y_j)$. The \textbf{Kantorovich optimal transport problem} \citep{peyre2020computationaloptimaltransport} between $\mu$ and $\nu$ is
\begin{equation}
  \mathrm{OT}_c(\mu, \nu) := \min_{P \in \Pi(\mu,\nu)} \langle P, C \rangle
  = \min_{P \in \Pi(\mu,\nu)} \sum_{i=1}^N \sum_{j=1}^M P_{ij}\, c(x_i, y_j).
  \label{eq:kantorovich-ot}
\end{equation}

The above objective function is a linear programming problem. Solving this directly is computationally expensive. Hence, to approximate the solution, an entropically regularized OT objective is usually preferred \citep{cuturi2013sinkhorn}, which replaces the linear program with a strictly convex problem:
\begin{align}
  \mathrm{OT}^{\chi}_{c}(\mu,\nu) := \min_{P\in \Pi(\mu,\nu)}\;\langle P,C\rangle - \chi H(P),\label{eq:ot-unreg}
\end{align}
where $H(P)=-\sum_{ij}P_{ij}\log P_{ij}$ and $\chi$ is the regularization parameter. This optimization problem is smooth and has a unique minimizer, which is computable via Sinkhorn iterations \citep{cuturi2013sinkhorn,peyre2020computationaloptimaltransport}. We explain the full details of regularized OT and the Sinkhorn Algorithm in \Cref{supp:ot}.

\yc{Also mention how we are different from Melynk, that they have a direct parametrization of reward with policy, where we only have access to samples and use empirical approximation of distribution i.e our approiach is more general where rewards are generated stochastically and we cannot use the reparameterization trick to express them directly as function of policy parameters}

\paragraph{Relation to FSD.}
We now describe how our FSD objective relates to the optimal transport objective. Using the results from \citet{Santambrogio2015OptimalTF} (Theorem 2.9 and Proposition 2.17) or \citet{melnyk2024distributional} (Theorem 1), for any two distributions $X$ and $Y$, the FSD objective can be framed as an optimal transport problem. Specifically,
\begin{align}\label{eq:otl}
  \mathcal{L}_\mathrm{FSD}(X,Y) = \mathrm{OT}_c(X,Y)
\end{align}
for the asymmetric convex cost function $c(x,y)=(y-x)_+$. We restate the relevant theorem in \Cref{supp:FSD} for completeness. We note that while \cite{melnyk2024distributional} also aims to do alignment via optimal transport, they apply stochastic dominance directly to reward distributions which are derived from a specific parametric form. On the other hand, we only have access to the cost distribution of a policy (independent from the reward distribution) which we access only through sampling. Hence, our framework is more general as we do not assume any specific parametric form.

\subsection{Spectral Risk Measures}

Spectral risk measures (SRMs) \citep{acerbi2002spectral} form a broad class of coherent \citep{artzner1999coherent}, law-invariant risk
functionals that aggregate quantile costs using a weighted spectrum over
confidence levels. They provide a flexible way to emphasize parts of the cost distribution (especially the tail) according to application-specific risk tolerance, rather than relying only on the mean. SRMs are widely used in fields like financial risk management and actuarial science to quantify and control exposure to adverse outcomes (\citet{dowd2008spectral,adam2008spectral}). Let $X$ be a real-valued random variable with quantile function
$Q_X : [0,1] \to \mathbb{R}$. A \emph{spectral risk measure} is defined as
\begin{equation}
  \rho_\phi(X)
  =
  \int_0^1 Q_X(q)\,\phi(q)\,dq,
\end{equation}
where $\phi : [0,1] \to \mathbb{R}_+$ is a non-negative, non-decreasing
weighting function satisfying
$\int_0^1 \phi(q)\,dq = 1.$
The function $\phi$ is referred to as the \emph{risk spectrum} and encodes
risk sensitivity across quantiles. Larger weights at higher quantiles emphasize higher
tail costs, while uniform weighting recovers the expectation.



%% file: sections/method.tex

We now present our safe alignment formulation -- \emph{Risk-sensitive Alignment via Dominance  (RAD)}. Standard Safe RLHF constrains the \emph{expected} cost, which controls only a single moment of the cost distribution.
RAD instead enforces a \emph{distributional} safety constraint using first-order stochastic dominance (FSD). We first introduce FSD as a principled way to ensure the learned policy's cost distribution is
stochastically smaller than a reference,
implemented via the asymmetric FSD surrogate $\mathcal{L}_{\mathrm{FSD}}$ (Eq.~\eqref{eq:fsd_loss}). We then show that reweighting quantiles of this surrogate recovers
Spectral Risk Measures (SRMs), yielding a unified framework in which the choice of weighting function $w(q)$ controls the risk profile of the learned policy.
Recall that our reward model is denoted by $r_{\phi}(x,y)$ and our cost model is denoted by $c_{\psi}(x,y)$ for input $x$ and ouptut $y$ and for learned parameters $\phi$ and $\psi$. For a policy $\pi$, let $C_\pi := c_\psi(x,y)$ denote the random variable of costs induced by sampling
$x\sim\mathcal{D}_x$ and $y\sim\pi(\cdot\mid x)$. Thus, we solve:
\begin{align}
  \max_\theta \quad & \mathbb{E}_{x\sim D_x,\,y\sim\pi_\theta(\cdot\mid x)}\!\Big[r_\phi(x,y)\Big]
  - \beta\,D_{\mathrm{KL}}\!\big(\pi_\theta(\cdot\mid x)\,\|\,\pi_{\mathrm{ref}}(\cdot\mid x)\big), \label{eq:sard_primal_obj}\\
  \text{s.t.}\quad & \mathcal{L}_{\mathrm{FSD}}\!\big(C_{\pi_\theta},\,C_{\pi_{\mathrm{ref}}}\big)\;\ge\;\kappa. \label{eq:sard_primal_constr}
\end{align}
Constraint~\eqref{eq:sard_primal_constr} encourages \emph{uniform} improvement across quantiles: since
$\mathcal{L}_{\mathrm{FSD}}(C_{\pi_\theta},\,C_{\pi_{\mathrm{ref}}})=\int_0^1 (Q_{C_{\pi_{\mathrm{ref}}}}(q)-Q_{C_{\pi_\theta}}(q))_+\,dq$,
larger values of $\mathcal{L}_{\mathrm{FSD}}\!\big(C_{\pi_\theta},C_{\pi_{\mathrm{ref}}}\big)$ correspond to larger positive gaps $Q_{C_{\pi_{\mathrm{ref}}}}(q)-Q_{C_{\pi_\theta}}(q)$ over $q\in[0,1]$.
Equivalently, it encourages $C_{\pi_{\mathrm{ref}}}\succeq_{\mathrm{FSD}} C_{\pi_\theta}$, i.e., the learned policy has stochastically smaller costs. 

To optimize our objective, we first absorb the KL term into the reward by defining the KL-regularized reward
$\tilde r(x,y)= r_\phi(x,y)-\beta\log \frac{\pi_\theta(y\mid x)}{\pi_{\mathrm{ref}}(y\mid x)}$ as done in \cite{dai2023safe},
and then use Dual Ascent \citep{gallier2019fundamentals} to optimize the Lagrangian relaxation of \eqref{eq:sard_primal_obj}--\eqref{eq:sard_primal_constr}:
\begin{equation}
  \max_\theta\min_{\lambda\ge 0}\;\underbrace{
    \mathbb{E}_{x\sim D_x,\,y\sim\pi_\theta(\cdot\mid x)}\!\Big[\tilde r(x,y)\Big]
  +\lambda\Big(\mathcal{L}_{\mathrm{FSD}}\!\big(C_{\pi_\theta},C_{\pi_{\mathrm{ref}}}\big)-\kappa\Big)}_{L(\theta,\lambda)}. \label{eq:sard_dual}
\end{equation}

\paragraph{Non-parametric (quantile-particle) representation of policy-induced cost distributions.}
\label{subsec:nonparam}
We now describe how we represent the cost distributions for our policies.
For a policy $\pi$, rather than assuming a parametric family for the cost distribution $C_\pi$, we use a non-parametric empirical approximation
via particles in the \emph{quantile space}. Fix quantile levels $\alpha_1,\dots,\alpha_N\in(0,1)$ and define the corresponding
cost-quantile particles:
\begin{equation}
  q_i(\pi) := Q_{C_\pi}(\alpha_i),\qquad i=1,\dots,N,
  \label{eq:quantile_particles}
\end{equation}
which we assume are ordered so that $q_1(\pi)\le \cdots \le q_N(\pi)$.
We approximate the cost distribution by the empirical measure:
\begin{equation}
  \hat\mu_\pi := \frac{1}{N}\sum_{i=1}^N \delta_{q_i(\pi)}.
  \label{eq:empirical_measure}
\end{equation}
In practice, $\hat\mu_\pi$ is obtained by sampling generations from $\pi$, evaluating their costs via $c_\psi$,
and computing empirical quantiles from the resulting samples; these quantiles serve as the particles defining the discrete
representation. Let $\hat{\mu}_{\pi_{\theta}}$ and $\hat{\mu}_{\pi_{\mathrm{ref}}}$ be the empirical measures for the cost distrivutions corresponding to policies $\pi_{\theta}$ and $\pi_{\mathrm{ref}}$ respctively. Then, the FSD objective is approximated as $\mathcal{L}_{\mathrm{FSD}}\!\big(C_{\pi_\theta},\,C_{\pi_{\mathrm{ref}}}\big) \approx \mathcal{L}_{\mathrm{FSD}}\!\big(\hat{\mu}_{\pi_{\theta}},\,\hat{\mu}_{\pi_{\mathrm{ref}}}\big)$. 
\paragraph{Computing the RAD policy gradients.}
The main technical challenge for computing the gradients of the RAD policy is differentiating the FSD term with respect to the policy parameters $\theta$.
To compute the gradients of the FSD objective effectively, first note that it can be interpreted as an optimal transport problem between the two cost distributions as described in~\eqref{eq:otl}. On adding entropic regularizion, the resulting optimization problem in~\eqref{eq:ot-unreg} is strictly convex and differentiable. Moreover, the entropic regularized optimal problem has smooth gradients and the unique minimizer of the objective can be computed efficiently using Sinkhorn iterations. Hence, we replace the FSD objective with the entropic regularzied FSD objective $\mathcal{L}^{\chi}_{\mathrm{FSD}}$ where $\chi$ is the regularization parameter (as in~\eqref{eq:ot-unreg}) \yc{$\beta$ already used in KL regularized rewards. Change this to $\chi$} \aj{We are also using the superscript for the weighted version of $\mathcal{L}$}. We next state the resulting policy-gradient estimator; the full derivation and implementation details are deferred to Supplementary Material ~\ref{app:sard_pg_derivation}.

\begin{theorem}[RAD policy-gradient estimator]\label{thm:sard_pg}
  Fix $\lambda\ge 0$ and define the dual objective $L(\theta,\lambda)$ by \eqref{eq:sard_dual}.
  Let $\alpha_1,\dots,\alpha_N\in(0,1)$ be fixed quantile levels and let
  $q_i(\pi_\theta):=Q_{C_{\pi_\theta}}(\alpha_i)$ denote the corresponding cost-quantile values. Using the empirical quantile-particle approximation from \Cref{eq:empirical_measure},
  the gradient $\nabla_\theta L(\theta,\lambda)$ admits the REINFORCE-form estimator
  \begin{equation}
    \nabla_\theta L(\theta,\lambda)
    =
    \mathbb{E}_{x\sim D_x,\,y\sim\pi_\theta(\cdot\mid x)}
    \Bigg[
      \Bigg(\tilde r(x,y)+\lambda\sum_{i=1}^N
        \Big(-\frac{\partial \mathcal{L}^{\chi}_{\mathrm{FSD}}}{\partial q_i(\pi_\theta)}\Big)\,
        \mathbf{1}\!\big(c_\psi(x,y)\le q_i(\pi_\theta)\big)
      \Bigg)
      \nabla_\theta\log\pi_\theta(y\mid x)
    \Bigg]. \label{eq:sard_pg}
  \end{equation}
  Moreover, for the weighted dominance objective $\mathcal{L}^{w}_{\mathrm{FSD}}$ (see \Cref{eq:weighted-fsd-loss-dfn}),
  the same estimator holds with an additional factor $w(\alpha_i)$ multiplying the $i$th summand. 
\end{theorem}
We optimize \eqref{eq:sard_dual} by alternating (i) ascent in $\theta$ using \eqref{eq:sard_pg} with REINFORCE \citep{williams1992simple} and variance-reduction scheme (RLOO) \citep{Kool2019Buy4R}, and (ii) Dual Ascent \citep{gallier2019fundamentals} in $\lambda$ to enforce the constraint.

\subsection{Universality of Weighted FSD for Controlling Spectral Risk Measures}\label{subsec:srm}


We now describe how we can change the FSD objective to represesnt various spectral risk measures. The FSD objective in Equation~\ref{eq:fsd_loss} weights all quantiles equally.
However, in many applications, tail performance is of primary concern, and it may be desirable to emphasize higher quantiles more heavily than lower ones.
To accommodate this, we introduce a quantile-weighted FSD objective with a nonnegative weighting function $w(q)$:
\begin{equation}
  \mathcal{L}_{\mathrm{FSD}}^{w}(X,Y)
  =
  \int_{0}^{1} w(q) \big(Q_{Y}(q) - Q_{X}(q)\big)_{+} dq,
  \label{eq:weighted-fsd-loss-dfn}
\end{equation}
where $w(q) \ge 0$ for all $q \in [0,1]$ and $\int_0^1 w(q) dq = 1$. While $\mathcal{L}_{\mathrm{FSD}}^{w}(X,Y)$ is not itself a spectral risk measure (SRM), it admits a direct structural relationship to the entire class of SRMs.
Recall that a spectral risk measure associated with weight function $w$ is defined as
\(
  \rho_w(Z) := \int_0^1 Q_Z(q)\, w(q)\, dq.
\)
Note that we have the following relationship:
\begin{equation}
  \int_{0}^{1} w(q) \big(Q_{Y}(q) - Q_{X}(q)\big) dq
  =
  \rho_w(Y) - \rho_w(X).
\end{equation}

Thus, weighted quantile differences correspond exactly to differences in spectral risk measures.
The following proposition formally states that the weighted FSD violations provide a decomposition of spectral risk differences.

\begin{proposition}
  Let $w : (0,1) \to \mathbb{R}_{+}$ be any nonnegative weight function with $\int_0^1 w(q) dq = 1$, and define
  \(
    \rho_w(Z) := \int_0^1 Q_Z(q)\, w(q)\, dq .
  \)
  For two random variables $X$ and $Y$, define
  {\small
    \[
      \mathcal{L}_{\mathrm{FSD}}^{w}(X,Y)
      :=
      \int_0^1 \big(Q_Y(q) - Q_X(q)\big)_+\, w(q)\, dq,
      \quad
      \mathcal{L}_{\mathrm{FSD}}^{w}(Y,X)
      :=
      \int_0^1 \big(Q_X(q) - Q_Y(q)\big)_+\, w(q)\, dq.
    \]
  }
  Then,
  {\small
    \[
      \rho_w(Y) - \rho_w(X)
      =
      \mathcal{L}_{\mathrm{FSD}}^{w}(X,Y)
      -
      \mathcal{L}_{\mathrm{FSD}}^{w}(Y,X).
      \implies
      - \mathcal{L}_{\mathrm{FSD}}^{w}(Y,X)
      \;\le\;
      \rho_w(Y) - \rho_w(X)
      \;\le\;
      \mathcal{L}_{\mathrm{FSD}}^{w}(X,Y).
    \]
  }
\end{proposition}

This proposition shows that differences in \emph{any} spectral risk measure can be decomposed into two weighted FSD violations.
Thus, weighted FSD provides a universal dominance-based control mechanism for the entire class of spectral risk measures.
\begin{corollary}
  If $\mathcal{L}_{\mathrm{FSD}}^w(X,Y) \geq \kappa$ for some $\kappa > 0$, then
  \[
    \rho_w(Y) - \rho_w(X)
    =
    \mathcal{L}_{\mathrm{FSD}}^w(X,Y)
    -
    \mathcal{L}_{\mathrm{FSD}}^w(Y,X)
    \ge
    \kappa
    -
    \mathcal{L}_{\mathrm{FSD}}^w(Y,X).
  \]
  In particular, if $Y \succeq_{\mathrm{FSD}} X$ so that
  $\mathcal{L}_{\mathrm{FSD}}^w(Y,X) = 0$, then
  \(
    \rho_w(X)
    \le
    \rho_w(Y) - \kappa.
  \)
  \label{cor:srm-fsd-relation}
\end{corollary}
Note that when $\kappa>\mathcal{L}_{\mathrm{FSD}}^w(Y,X)$, enforcing $\mathcal{L}_{\mathrm{FSD}}^w(X,Y) \geq \kappa$ implies reduction in $\rho_w(X)$ w.r.t. $\rho_w(Y)$.
This corollary formalizes the universality property:
enforcing a lower bound on the weighted FSD violation guarantees improvement under the corresponding spectral risk measure, provided reverse violations vanish.

Substituting the cost distributions $X = C_{\pi_{\theta}}$ and
$Y = C_{\pi_{\mathrm{ref}}}$ into Corollary~\ref{cor:srm-fsd-relation}, enforcing
\[
  \mathcal{L}_{\mathrm{FSD}}^w(C_{\pi_{\theta}},C_{\pi_{\mathrm{ref}}}) \ge \kappa
  \implies
  \rho_w\!\left(C_{\pi_{\theta}}\right)
  \le
  \rho_w\!\left(C_{\pi_{\mathrm{ref}}}\right)
  -
  \kappa
  +
  \mathcal{L}_{\mathrm{FSD}}^w(C_{\pi_{\mathrm{ref}}},C_{\pi_{\theta}}).
\]

In particular, if
\( C_{\pi_{\mathrm{ref}}}\succeq_{\mathrm{FSD}} C_{\pi_{\theta}} \),
then \( \mathcal{L}_{\mathrm{FSD}}^w(C_{\pi_{\mathrm{ref}}},C_{\pi_{\theta}})=0 \), yielding the strict spectral risk improvement
\(
  \rho_w\!\left(C_{\pi_{\theta}}\right)
  \le
  \rho_w\!\left(C_{\pi_{\mathrm{ref}}}\right)
  -
  \kappa.
\)
In practice, as optimization proceeds under the weighted FSD constraint,
\( \mathcal{L}_{\mathrm{FSD}}^w(C_{\pi_{\theta}},C_{\pi_{\mathrm{ref}}}) \) increases while
\( \mathcal{L}_{\mathrm{FSD}}^w(C_{\pi_{\textrm
{ref}}},C_{\pi_{\theta}}) \) decreases, typically approaching zero near convergence.

Table~\ref{tab:spectral_weights} lists representative spectral risk measures and their associated quantile-weighting functions. Figure ~\ref{fig:srm-illustration} illustrates the spectral weighting function for different SRMs considered in Table~\ref{tab:spectral_weights}. For example, if $w(q)$ corresponds to $\mathrm{CVaR}_{\alpha}$, then enforcing
\(
  \mathcal{L}_{\mathrm{FSD}}^w(C_{\pi_{\theta}},C_{\pi_{\mathrm{ref}}}) \ge \kappa
\)
guarantees
\(
  \mathrm{CVaR}_{\alpha}\!\left(C_{\pi_{\theta}}\right)
  \le
  \mathrm{CVaR}_{\alpha}\!\left(C_{\pi_{\mathrm{ref}}}\right)
  -
  \kappa,
\)
whenever reverse violations vanish.
If $w(q)$ is uniform, corresponding to the unweighted FSD objective, then enforcing the FSD constraint guarantees that, at convergence,
\(
  \mathbb{E}[C_{\pi_{\theta}}]
  \le
  \mathbb{E}[C_{\pi_{\mathrm{ref}}}]
  -
  \kappa.
\)

\begin{table}[t]
  \centering
  \small







  {\footnotesize
    \begin{tabular}{l c c c c c c c}
      \hline
      \textbf{Risk} & \textbf{Mean} & \textbf{VaR$_\alpha$\footnotemark} & \textbf{CVaR$_\alpha$} & \textbf{Linear} & \textbf{Exponential} & \textbf{Power} & \textbf{Wang} \\
      \textbf{Measure}
      &
      & \textbf{}
      & \textbf{}
      & \textbf{Spectral}
      & \textbf{Spectral}
      & \textbf{Spectral}
      & \textbf{Distortion}
      \\
      \hline

      \textbf{$\mathbf{w(q)}$}
      & $1$
      & $\delta(q-\alpha)$
      & $\frac{1}{1-\alpha}\mathbf{1}_{[\alpha,1]}(q)$
      & $2q$
      & $\frac{\lambda e^{\lambda q}}{e^\lambda - 1}$
      & $(1+\lambda) q^\lambda$
      & $\frac{\Phi(\Phi^{-1}(q)+\lambda)}{1-\Phi(\lambda)}$
      \\
      \hline
    \end{tabular}
  }
  \caption{Examples of weight functions $w(q)$ that induce corresponding spectral risk measure constraints when used in $\mathcal{L}_{\mathrm{FSD}}^{w}$. $\lambda > 0$ denotes a risk aversion parameter and $\Phi$ in the Wang Distortion \citep{wang1996premium} risk measure denotes the CDF of a standard normal distribution.}
  \label{tab:spectral_weights}
\end{table}

\footnotetext{VaR is not a spectral risk measure as it is not a coherent risk measure, but can still be represented in the RAD framework.}

Thus, though a weighted FSD is not a spectral risk measure by itself, it serves as a universal dominance-based control mechanism for the entire class of spectral risk measures.

%% file: sections/experiments.tex
We investigate the following research questions: [Q1] \emph{How do models aligned using RAD compare to methods that enforce safety via expected cost constraints, in terms of helpfulness and harmfulness?} [Q2] \emph{How do RAD-aligned models perform against baselines wrt harmlessness of responses, on out-of-distribution data?}

We follow the standard RLHF pipeline for alignment. As our base model, we use Qwen2.5-3B \citep{qwen2025qwen25technicalreport}. We first fine-tune this model on the Alpaca dataset \citep{alpaca} \raj{citation?} to obtain the SFT initialization. For reward and cost modeling, we use the BeaverTails dataset~\citep{ji2023beavertailsimprovedsafetyalignment}, which provides pairwise preference annotations along two dimensions: helpfulness and harmfulness. The helpfulness preferences are used to train a reward model, while the harmfulness preferences are used to train a cost model. Both models are trained using the standard Bradley-Terry objective \citep{bradley1952rank} in \Cref{eq:bt_rm_loss}. Importantly, we adopt the same fine-tuning and reward/cost modeling procedure used in Safe RLHF~\citep{dai2023safe}, ensuring a controlled comparison.

In the RL stage, RAD employs a REINFORCE-based policy gradient method, using \Cref{eq:sard_pg}, with RLOO (Leave-One-Out) baseline and two sampled responses per prompt ($k = 2$). Additional implementation
details, including hyperparameters, are provided in supplementary material \ref{supp:implementation-details}. We set $\kappa=10$ for our RAD optimization objective in 
\Cref{eq:sard_primal_constr}. To provide a fair comparison with the baselines, we set the Safe-RLHF cost threshold $\tau=-10$ in \Cref{eq:safe-rlhf-obj}, following Corollary~\ref{cor:srm-fsd-relation}, since we observed $\mathbb{E}_{x \sim \mathcal{D}, y \sim 
\pi_{\textrm{ref}}(\cdot\vert x)}[c_{\psi}(x,y)] \approx 0$.

\subsection{Model Evaluations}

We compare models aligned using RAD and Safe RLHF using the same reward and cost models trained on BeaverTails preferences~\citep{ji2023beavertailsimprovedsafetyalignment}. Since RAD is a dominance based alignment framework rather than a single objective, we train multiple models corresponding to different spectral weighting schemes over the quantiles in the FSD objective. These weighting schemes are listed in Table~\ref{tab:spectral_weights}.

We evaluate the helpfulness and harmlessness of all model responses, using the prompts in the test set of BeaverTails. We adopt a pairwise evaluation protocol in which we designate one model as the \emph{blue} model and another as the \emph{red} model. Typically, the \emph{red} model represents a baseline (Safe RLHF or SFT) and the \emph{blue} model represents a RAD variant, though this assignment can vary across comparisons.

\paragraph{Harmlessness}

We use the trained cost model to judge the harmlessness of model generations for all considered approaches. From Table~\ref{tab:safe-response-prop}, RAD models achieve a higher proportion of safe responses compared to both baselines -- SFT and
Safe RLHF -- highlighting the benefits of enforcing distributional dominance constraints over expected cost alone.

\begin{table}[t]
  \centering
  {\footnotesize
    \begin{tabular}{>{\centering\arraybackslash}p{4cm} >{\centering\arraybackslash}p{2.5cm} >{\centering\arraybackslash}p{2.5cm} >{\centering\arraybackslash}p{1.5cm} >{\centering\arraybackslash}p{1.5cm}}
      \toprule
      \makecell{Competition \\ (Red vs Blue)} & 
        \makecell{Reward of Blue \\ Higher} & 
        \makecell{Reward of Blue \\ Lower} & 
        \makecell{Overall \\ Blue} & 
        \makecell{Overall \\ Red} \\
      \midrule
      safe-rlhf vs fsd-uniform             & $0.96\pm0.01$ & $0.83\pm0.14$ & $0.92\pm0.05$ & \underline{\textbf{$0.93\pm0.00$}} \\
      safe-rlhf vs fsd-var                 & $0.88\pm0.10$ & $0.88\pm0.10$ & $0.88\pm0.10$ & \underline{\textbf{$0.93\pm0.00$}} \\
      safe-rlhf vs fsd-cvar                & $0.96\pm0.00$ & $0.98\pm0.00$ & \underline{\textbf{$0.97\pm0.00$}} & $0.93\pm0.00$ \\
      safe-rlhf vs fsd-spectral-linear     & $0.97\pm0.00$ & $0.93\pm0.04$ & \underline{\textbf{$0.94\pm0.03$}} & $0.93\pm0.00$ \\
      safe-rlhf vs fsd-spectral-wang       & $0.97\pm0.01$ & $0.95\pm0.00$ & \underline{\textbf{$0.96\pm0.00$}} & $0.93\pm0.00$ \\
      safe-rlhf vs fsd-spectral-power      & $0.97\pm0.01$ & $0.94\pm0.02$ & \underline{\textbf{$0.96\pm0.01$}} & $0.93\pm0.00$ \\
      safe-rlhf vs fsd-spectral-exponential& $0.98\pm0.01$ & $0.94\pm0.03$ & \underline{\textbf{$0.96\pm0.01$}} & $0.93\pm0.00$ \\
      \midrule
      sft vs fsd-uniform                   & $0.95\pm0.03$ & $0.65\pm0.16$ & \underline{\textbf{$0.92\pm0.05$}} & $0.55\pm0.00$ \\
      sft vs fsd-var                       & $0.93\pm0.06$ & $0.82\pm0.10$ & \underline{\textbf{$0.88\pm0.10$}} & $0.55\pm0.00$ \\
      sft vs fsd-cvar                      & $0.98\pm0.00$ & $0.92\pm0.01$ & \underline{\textbf{$0.97\pm0.01$}} & $0.55\pm0.00$ \\
      sft vs fsd-spectral-linear           & $0.97\pm0.01$ & $0.74\pm0.15$ & \underline{\textbf{$0.94\pm0.03$}} & $0.55\pm0.00$ \\
      sft vs fsd-spectral-wang             & $0.97\pm0.00$ & $0.75\pm0.03$ & \underline{\textbf{$0.96\pm0.00$}} & $0.55\pm0.00$ \\
      sft vs fsd-spectral-power            & $0.97\pm0.01$ & $0.76\pm0.08$ & \underline{\textbf{$0.95\pm0.01$}} & $0.55\pm0.00$ \\
      sft vs fsd-spectral-exponential      & $0.98\pm0.01$ & $0.66\pm0.17$ & \underline{\textbf{$0.96\pm0.01$}} & $0.55\pm0.00$ \\
      \midrule
      sft vs safe-rlhf                     & $0.96\pm0.00$ & $0.61\pm0.03$ & \underline{\textbf{$0.93\pm0.00$}} & $0.55\pm0.00$ \\
      \bottomrule
    \end{tabular}
  }
  \caption{
    Comparison of the proportion of safe responses produced by the blue model, as judged by the cost model, across competing model pairs. We further decompose this proportion into two subsets based on whether the blue model's response achieved higher reward than the red model's response (first two columns). The ``Overall Blue'' column reports the overall proportion of safe responses for the blue model, while the ``Overall Red'' column reports the same for the red model. We observe that RAD-aligned models consistently produce a higher proportion of safe responses than the baselines, as evidenced by the comparison of the final two columns. Values are reported as mean $\pm$ standard deviation across 3 random seeds.
  }
  \label{tab:safe-response-prop}
\end{table}

Let $C_{\pi_{\textrm{blue}}}$ and $C_{\pi_{\textrm{red}}}$ denote the random variables corresponding to the cost distributions induced by the blue and red models, respectively. We define the dominance difference as
\[
  D_{\textrm{FSD}}^{w}(C_{\pi_{\textrm{blue}}}, C_{\pi_{\textrm{red}}}) :=
  \mathcal{L}_{\textrm{FSD}}^{w}(C_{\pi_{\textrm{blue}}}, C_{\pi_{\textrm{red}}})
  - \mathcal{L}_{\textrm{FSD}}^{w}(C_{\pi_{\textrm{red}}}, C_{\pi_{\textrm{blue}}}).
\]

Intuitively, for the blue model to have a safer cost distribution than the red model, we would want
$\mathcal{L}_{\textrm{FSD}}^{w}(C_{\pi_{\textrm{blue}}}, C_{\pi_{\textrm{red}}})$ to be high and
$\mathcal{L}_{\textrm{FSD}}^{w}(C_{\pi_{\textrm{red}}}, C_{\pi_{\textrm{blue}}})$ to be low.
This implies a high positive $D_{\textrm{FSD}}^{w}(C_{\pi_{\textrm{blue}}}, C_{\pi_{\textrm{red}}})$.

We compare the competing models using $D_{\textrm{FSD}}^{w}(C_{\pi_{\textrm{blue}}}, C_{\pi_{\textrm{red}}})$, reporting both the weighted and unweighted dominance difference. The weighted dominance difference corresponds to the difference in spectral risk measures under the chosen weighting: a positive value indicates a reduction in the weighted spectral risk measure for the blue model compared to the red model. The unweighted dominance difference corresponds to the difference in average cost: a positive value indicates a lower average cost for the blue model relative to the red model. Table~\ref{tab:cost-dominance} presents the dominance results.
We use an unnormalized weighting in the Dominance Difference; this does not affect the results, as all values are scaled by a constant factor for a fixed weighting.

From Table~\ref{tab:cost-dominance}, we observe that most RAD models, across different quantile weighting functions, achieve a positive weighted dominance metric relative to the baselines, implying a  reduction in the corresponding spectral risk measure. Together with the higher proportion of safe responses, these results demonstrate the benefits of enforcing distributional dominance constraints over
methods that constrain only a central statistic of the cost distribution, such as the expected cost. Importantly, RAD's improvements are consistent across multiple weighting functions, suggesting that its effectiveness is not tied to any particular risk
metric. Practitioners can therefore select the weighting function that best reflects their deployment context and risk tolerance --
for instance, preferring CVaR-style weightings in high-stakes settings or uniform weighting when average cost reduction is the primary objective.

\begin{table}[t]
  \centering
  {\footnotesize
    \begin{tabular}{lcc}
      \toprule
      Competetion (Red vs Blue) & $D_{\textrm{FSD}}(C_{\pi_{\textrm{blue}}}, C_{\pi_{\textrm{red}}})$ & $D_{\textrm{FSD}}^{w}(C_{\pi_{\textrm{blue}}}, C_{\pi_{\textrm{red}}})$ \\
      \midrule
      safe-rlhf vs fsd-uniform & $+7.92 \pm 11.23$ & \underline{\textbf{$+7.92 \pm 11.23$}} \\
      safe-rlhf vs fsd-var & $-66.06 \pm 46.66$ & $-2.95 \pm 10.93$ \\
      safe-rlhf vs fsd-cvar & $-12.54 \pm 2.80$ & \underline{\textbf{$+56.61 \pm 10.39$}} \\
      safe-rlhf vs fsd-spectral-linear & $-7.33 \pm 7.63$ & $-0.49 \pm 11.59$ \\
      safe-rlhf vs fsd-spectral-wang & $+5.74 \pm 5.00$ & \underline{\textbf{$+27.91 \pm 14.99$}} \\
      safe-rlhf vs fsd-spectral-power & $-3.50 \pm 6.34$ & \underline{\textbf{$+9.45 \pm 8.67$}} \\
      safe-rlhf vs fsd-spectral-exponential & $+8.18 \pm 11.04$ & \underline{\textbf{$+22.16 \pm 10.04$}} \\
      \midrule
      sft vs fsd-uniform & $+147.12 \pm 12.43$ & \underline{\textbf{$+147.12 \pm 12.43$}} \\
      sft vs fsd-var & $+73.06 \pm 46.52$ & \underline{\textbf{$+17.59 \pm 10.83$}} \\
      sft vs fsd-cvar & $+126.84 \pm 1.68$ & \underline{\textbf{$+139.83 \pm 7.52$}} \\
      sft vs fsd-spectral-linear & $+131.24 \pm 11.22$ & \underline{\textbf{$+135.64 \pm 15.54$}} \\
      sft vs fsd-spectral-wang & $+144.63 \pm 1.56$ & \underline{\textbf{$+432.69 \pm 4.63$}} \\
      sft vs fsd-spectral-power & $+135.91 \pm 5.54$ & \underline{\textbf{$+134.40 \pm 3.95$}} \\
      sft vs fsd-spectral-exponential & $+147.88 \pm 9.84$ & \underline{\textbf{$+145.47 \pm 9.59$}} \\
      \midrule
      sft vs safe-rlhf & $+138.51 \pm 2.45$ & - \\
      \bottomrule
    \end{tabular}
  }
  \caption{
    Unweighted and weighted dominance differences for all pairs of competing models.
    A positive weighted dominance difference for the blue model indicates a reduction in the corresponding spectral risk measure relative to the red model. The weighted dominance difference uses the weighting corresponding to the blue model. Some weighting schemes (e.g., CVaR and VaR) exhibit negative unweighted dominance differences but positive weighted dominance differences.
    This occurs because CVaR only weights the higher quantiles while setting others to zero, and VaR is a Dirac mass at a particular quantile (Gaussian-smoothed for implementation purposes), disregarding other quantiles. Values are reported as mean $\pm$ standard deviation across 3 random seeds.
  }
  \label{tab:cost-dominance}
\end{table}

\paragraph{Helpfulness}

\begin{figure}  
  \centering
  \includegraphics[width=0.9\textwidth]{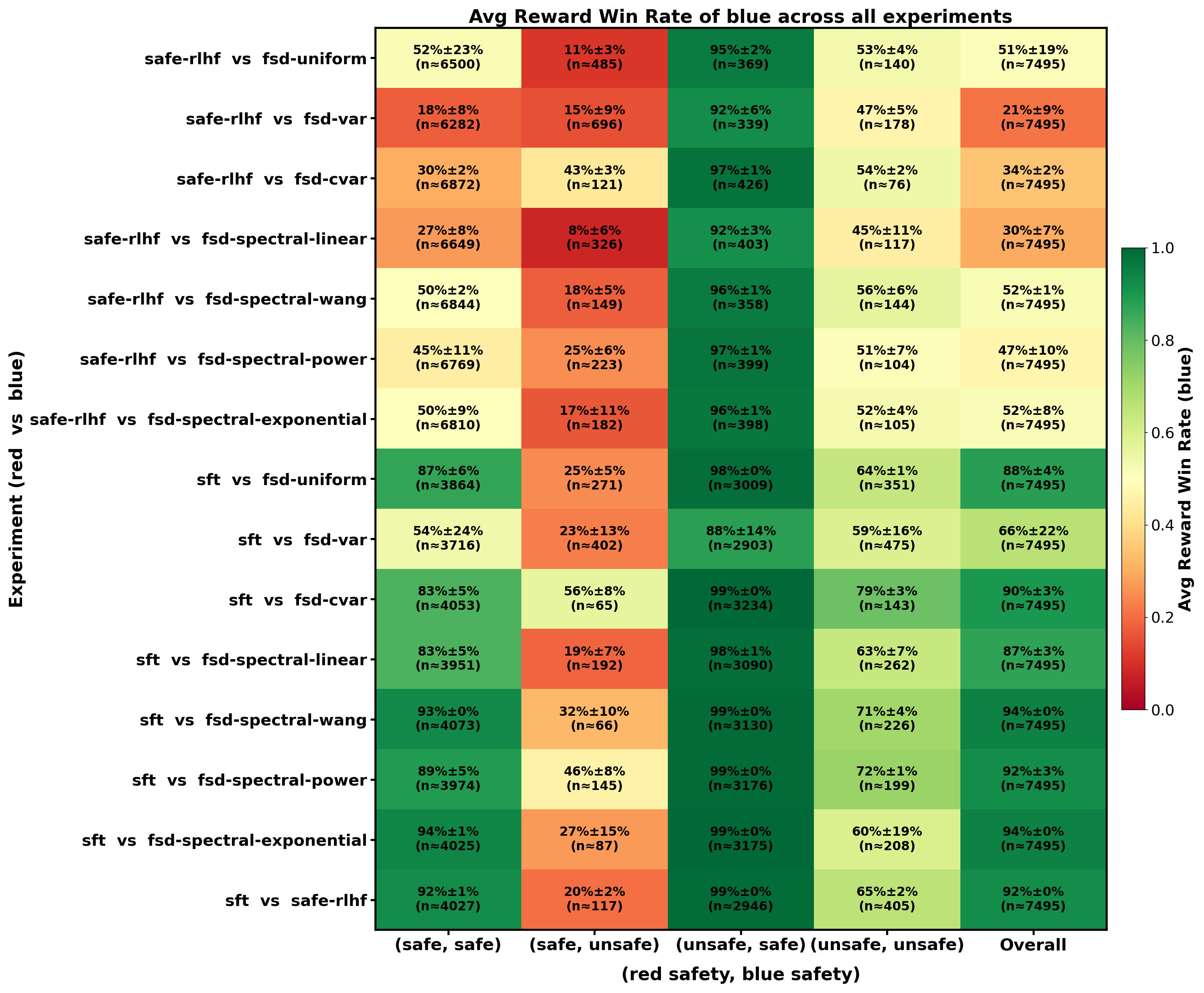}
  \caption{
     Average reward win rates between competing models, across 3 seeds.
  }
  \label{fig:reward-winrates-test}
\end{figure}

We use the trained reward model, to judge the helpfulness of model generations for all the approaches we consider. We compute the reward win-rates, pitting two models against each other, with their response helpfulness judged by the reward model, which assigns a higher score to more helpful responses. The results are shown in Figure \ref{fig:reward-winrates-test}. In each cell, in addition to win rate, we also provide the count of the number of responses compared. Each row corresponds to a pairwise competition (red vs. blue). We report the reward win rate of the blue model, defined as the percentage of prompts for which the blue model's response achieves higher reward than the red model's response. The x-axis partitions results by safety outcome combinations of the two models: (safe, safe), (safe, unsafe), (unsafe, safe), and (unsafe, unsafe), where the first entry corresponds to the red model and the second to the blue model. For example, (safe, unsafe) denotes the subset of prompts where the red model produces a safe response and the blue model produces an unsafe response. The final column reports the overall win rate across all prompts. 

From Figure~\ref{fig:reward-winrates-test}, RAD-aligned models achieve higher rewards than the SFT baseline, as evidenced by consistently high reward win rates. Against Safe-RLHF, FSD-VaR, FSD-CVaR, and FSD-Spectral-Linear obtain lower reward win rates, suggesting these weighting schemes are more risk-averse at the expense of helpfulness -- a tradeoff that may be desirable in risk-sensitive deployment contexts such as medical or legal applications. The remaining variants achieve comparable reward win rates against Safe-RLHF, indicating helpfulness parity.



\paragraph{Summary}
From the results on the harmlessness and helpfulness of model responses, we observe that RAD-aligned methods produce a higher proportion of safe
responses compared to the baselines, and exhibit a higher positive weighted dominance difference, implying a reduction in spectral risk relative to the baselines. Furthermore, for spectral weighting schemes -- specifically Spectral-Wang, Spectral-Power, Spectral-Exponential, and uniform
-- these gains in safety are made while maintaining parity in helpfulness with Safe RLHF.

\subsection{Out-of-Distribution Harmlessness: HarmBench Evaluation}

To assess generalization beyond the training distribution, we evaluate
all models on HarmBench \citep{mazeika2024harmbenchstandardizedevaluationframework}, an out-of-distribution
red-teaming benchmark comprising adversarially constructed harmful prompts
across diverse risk categories. Critically, neither the reward model, cost
model, nor any model checkpoint was trained or selected using HarmBench
data, making this a clean held-out evaluation.

We use GPT-4o-mini \citep{openai2024gpt4omini} as an independent pairwise harmlessness judge,
following standard LLM-as-a-judge \citep{zheng2023judgingllmasajudgemtbenchchatbot} evaluation protocol \cite{}. For each
prompt, the judge compares responses from two models and determines which
is less harmful. We provide the GPT evaluation results in \Cref{tab:harmbench-gpt-eval}

All RAD variants substantially outperform the SFT baseline, confirming that our constrained RL training, with the SFT policy as the reference, improves harmlessness and generalizes to unseen prompts.  Against Safe-RLHF, spectral variants -- Spectral-Power, Spectral-Exponential, Spectral-Linear, CVaR, and VaR -- achieve favorable win-loss ratios, suggesting that upweighting the tail of the cost distribution confers greater robustness to distributional shift. 



\begin{table}[t]
  \centering
  {\footnotesize
    \begin{tabular}{lcc}
      \toprule
      Model Name & SFT & Safe RLHF \\
      \midrule
      fsd-uniform & \underline{\textbf{67.1\%}}{\scriptsize$\pm$2.1} / 7.8\%{\scriptsize$\pm$4.2} / 25.1\%{\scriptsize$\pm$6.3} & 38.0\%{\scriptsize$\pm$6.9} / 22.2\%{\scriptsize$\pm$15.0} / \underline{\textbf{39.8\%}}{\scriptsize$\pm$8.4} \\
      fsd-var & \underline{\textbf{76.6\%}}{\scriptsize$\pm$3.7} / 10.6\%{\scriptsize$\pm$2.7} / 12.7\%{\scriptsize$\pm$2.3} & \underline{\textbf{39.6\%}}{\scriptsize$\pm$10.6} / 28.6\%{\scriptsize$\pm$16.8} / 31.8\%{\scriptsize$\pm$9.7} \\
      fsd-cvar & \underline{\textbf{72.9\%}}{\scriptsize$\pm$2.4} / 15.7\%{\scriptsize$\pm$1.3} / 11.4\%{\scriptsize$\pm$2.4} & \underline{\textbf{28.6\%}}{\scriptsize$\pm$7.8} / 46.5\%{\scriptsize$\pm$2.1} / 24.9\%{\scriptsize$\pm$6.6} \\
      fsd-spectral-linear & \underline{\textbf{75.3\%}}{\scriptsize$\pm$6.3} / 12.3\%{\scriptsize$\pm$2.0} / 12.4\%{\scriptsize$\pm$5.1} & \underline{\textbf{34.2\%}}{\scriptsize$\pm$2.8} / 37.0\%{\scriptsize$\pm$12.8} / 28.9\%{\scriptsize$\pm$13.4} \\
      fsd-spectral-wang & \underline{\textbf{67.3\%}}{\scriptsize$\pm$7.9} / 13.4\%{\scriptsize$\pm$0.6} / 19.3\%{\scriptsize$\pm$7.9} & 24.7\%{\scriptsize$\pm$1.7} / 49.7\%{\scriptsize$\pm$7.4} / \underline{\textbf{25.5\%}}{\scriptsize$\pm$8.1} \\
      fsd-spectral-exponential & \underline{\textbf{74.7\%}}{\scriptsize$\pm$1.6} / 12.6\%{\scriptsize$\pm$1.2} / 12.7\%{\scriptsize$\pm$1.6} & \underline{\textbf{28.0\%}}{\scriptsize$\pm$6.1} / 52.2\%{\scriptsize$\pm$7.0} / 19.9\%{\scriptsize$\pm$2.2} \\
      fsd-spectral-power & \underline{\textbf{71.0\%}}{\scriptsize$\pm$2.7} / 14.8\%{\scriptsize$\pm$2.1} / 14.1\%{\scriptsize$\pm$4.1} & \underline{\textbf{30.4\%}}{\scriptsize$\pm$2.0} / 48.9\%{\scriptsize$\pm$1.1} / 20.7\%{\scriptsize$\pm$2.7} \\
      \midrule
      Safe RLHF & \underline{\textbf{68.1\%}}{\scriptsize$\pm$2.6} / 20.4\%{\scriptsize$\pm$1.1} / 11.5\%{\scriptsize$\pm$2.5} & - \\
      \bottomrule
    \end{tabular}
  }
  \caption{
    Pairwise harmlessness evaluation on HarmBench using GPT-4o-mini as an independent judge. Each cell reports Win / Tie / Lose rates of the row model against the column model (mean $\pm$ std over 3 seeds). The dominant outcome -- win rate if the row model wins, loss rate if it loses -- is bold underlined. All FSD variants substantially outperform the SFT baseline. Against Safe RLHF, variants that upweight the tail (Eg: Exponential, Power, Linear, CVaR, and Var) achieve favorable win-loss ratios, demonstrating improved harmlessness generalization to out-of-distribution prompts.
  }
  \label{tab:harmbench-gpt-eval}
\end{table}

%% file: sections/relatedwork.tex
\aj{Added two more lines}
Standard RLHF typically optimizes LLM behavior using a single preference-based reward signal, which couples helpfulness and harmlessness and can yield either unsafe completions or unhelpful blanket refusals in adversarial or sensitive prompts \citep{ouyang2022training,bai2022constitutional}. Complementary mitigation strategies aim to reduce harmful outputs using safety critics, filtering mechanisms, and curated datasets \citep{xu2020recipes,thoppilan2022lamda,ziegler2022adversarial}.
To make the objectives more controllable, later approaches separate helpfulness and harmlessness signals, e.g., by combining distinct model scores or by using safety as an explicit optimization constraint \citep{glaese2022improving,mu2024rule,touvron2023llama,ji2023beavertailsimprovedsafetyalignment}.
Safe RLHF makes this trade-off explicit by training separate reward and cost models and then maximizing reward under an expected-cost constraint in a constrained-MDP view \citep{dai2023safe,altman2021constrained}. HC-RLHF further addresses statistical uncertainty by instantiating this constraint within the Seldonian framework and adding a separate safety-screening stage based on upper-confidence bounds computed from held-out data \citep{thomas2019preventing,chittepu2025reinforcement}. 

In parallel, a growing line of work argues that expectation objectives can overlook distributional desiderata, advocating stochastic dominance as an explicit ordering criterion and, in some cases, leveraging optimal transport (OT) formulations for tractable learning and matching \citep{melnyk2024distributional,farajzadeh2025imitation,cen2024beyond}. The closest complementary advances in stochastic optimization study how to enforce stochastic-dominance constraints via dual/Lagrangian characterizations and surrogate methods to make dominance-constrained optimization computationally practical \citep{dentcheva2003optimization,dai2023learning,cen2024beyond}. Building on both the constrained-RLHF and dominance-based optimization perspectives, our work adopts the safe-RLHF framework but replaces expectation-based safety control with a first-order stochastic dominance constraint on the entire cost distribution relative to a reference policy, leveraging OT-based relaxations to render the dominance constraint differentiable and optimizable. Unlike prior Safe RLHF methods that impose scalar (possibly high-confidence) expected-cost constraints \citep{dai2023safe,chittepu2025reinforcement}, our approach enforces an OT-optimized FSD constraint on the full cost distribution, thereby targeting distributional safety guarantees beyond average-cost control.

%% file: sections/conclusion.tex
In this work, we argued that safety in RLHF should control the full distribution of policy-induced costs, rather than only a central statistic, like their expectation. We proposed \emph{Risk-sensitive Alignment via Dominance} (RAD), which enforces first-order stochastic dominance relative to a reference policy and admits a practical optimization procedure via quantile surrogates and entropic optimal transport.
By introducing quantile-weighted objectives, RAD also provides a unified mechanism for controlling a broad class of spectral risk measures, allowing safety preferences to be tuned toward different parts of the cost distribution.
Empirically, our results show that RAD can improve harmlessness over expectation-based baselines while remaining competitive in helpfulness, with several variants also showing stronger robustness on out-of-distribution harmfulness evaluations.

%% file: sections/supplementary.tex
\section{Implementation Details}
\label{supp:implementation-details}

We use the \href{https://github.com/PKU-Alignment/safe-rlhf}{Safe-RLHF repository} \citep{dai2023safe} and build RAD on top of their publicly available codebase. For our Safe-RLHF runs, we use the hyperparameters 
reported in their paper. For our RAD implementation, we use REINFORCE \citep{williams1992simple} with the RLOO variance-reduction baseline \citep{Kool2019Buy4R} and $k=2$ samples per prompt. To compute the quantiles required for the RAD policy gradient, we gather samples 
across all GPUs before computing quantiles, increasing the number of available samples. The particle gradient is computed using the Python Optimal Transport library \citep{flamary2021pot}. All RAD models were 
trained on two NVIDIA A100 GPUs, while Safe-RLHF runs required four NVIDIA A100 GPUs due to the additional memory overhead of storing and training the critic network.

\begin{table}[h]
\centering
\begin{tabular}{lc}
\toprule
Hyperparameter & Value \\
\midrule
Sinkhorn regularization $\chi$ & $0.01$ \\
KL coefficient $\beta$ (same value used for Safe RLHF) & $0.1$ \\
$\kappa$ & $10$ \\
Spectral-Wang $\lambda$ & $0.7$ \\
Spectral-Exponential $\lambda$  & $3.0$ \\
Spectral-Power $\lambda$ & $2.0$ \\
Gaussian bandwidth (for VaR) & $0.1$ \\
$\alpha$ ($\textrm{CVaR}_{\alpha}$ / $\textrm{VaR}_{\alpha}$) & $0.9$ \\
\bottomrule
\end{tabular}
\caption{RAD-specific hyperparameters. All other hyperparameters are the same as ones used in Safe-RLHF, unless specified otherwise.}
\label{tab:hyperparams}
\end{table}

\section{RLHF}
\label{supp:RLHF}

Reinforcement Learning from Human Feedback (RLHF) \citep{Christiano2017DeepRL,Ouyang2022TrainingLM} is currently the predominant strategy for aligning Large Language Models (LLMs) with human intent and preferences. The process typically begins with a pre-trained base model, which has been trained on internet-scale data via a next-token prediction objective. The standard RLHF pipeline generally consists of three distinct stages: Supervised Fine-Tuning (SFT), Reward Modeling (RM), and Reinforcement Learning (RL). We detail each of these stages below.

\paragraph{Supervised Fine-Tuning} In the SFT stage, the pre-trained model is trained to follow instructions, via a next-token prediction objective. This process utlizes a high quality dataset of prompt-responses pairs $D_\mathrm{sft}$, where the responses are provided by either a human or LLM annotator \citep{Bai2022ConstitutionalAH}. We refer to the resulting policy from the SFT stage as $\pi_\mathrm{sft}$.

\paragraph{Reward Modeling.}
In the reward modeling stage, we train a reward model to capture human preferences. This stage relies on a dataset of human preferences $D_{\mathrm{pref}} = \{(x_i, y_i^+, y_i^-)\}_{i=1}^N$, where $x_i$ is a prompt, $y_i^+$ is the preferred response to $x_i$, and $y_i^-$ is the dispreferred response to $x_i$. Preference annotations are collected either from human annotators or LLM annotators. Preferences are modeled using the Bradley--Terry preference model \citep{bradley1952rank}, where the log-odds of an observed preference equals the difference between the rewards assigned to the preferred and dispreferred responses by a latent reward function $r(x,y)$:
\begin{equation}
  P(y^+ \succ y^- \mid x)
  =
  \frac{e^{r(x,y^+)}}{e^{r(x,y^+)} + e^{r(x,y^-)}}
  =
  \sigma\big(r(x,y^+) - r(x,y^-)\big),
\end{equation}
where $\sigma$ denotes the sigmoid function. A parameterized reward function with parameters $\phi$ is learned to approximate the latent reward function through maximum likelihood estimation on the preference dataset $D_{\mathrm{pref}}$, using the following objective:
\begin{equation}
  \min_{\phi}
  -\mathbb{E}_{(x,y^{+},y^{-}) \sim D_\mathrm{pref}}
  \big[
    \log \sigma\big(r_{\phi}(x,y^+) - r_{\phi}(x,y^-)\big)
  \big].
  \label{eq:bt_rm_loss}
\end{equation}

\paragraph{Reinforcement Learning} In the Reinforcement Learning stage, a language model or policy is trained to generate responses that are preferred by humans by maximizing the reward of generated responses, as measured by the reward model $r_{\phi}(x,y)$. However, directly optimizing the reward of model generations can lead to degradation in response quality \citep{stiennon2022learningsummarizehumanfeedback,Ouyang2022TrainingLM, Jaques2019WayOB} due to a phenomenon known as reward overoptimization \citep{Gao2022ScalingLF,Rafailov2024ScalingLF}, where the model overfits to imperfections in the learned proxy reward function. To mitigate this effect, a KL penalty term is added to the objective to penalize the model (policy) from drifting too far away from a reference policy, usually chosen to be $\pi_{\mathrm{sft}}$. The RL objective can be expressed as

\begin{equation}
  \max_{\theta} \mathbb{E}_{x \sim \mathcal{D}_x, y \sim \pi_{\theta}(.\vert x)} [r_{\phi}(x,y)] - \beta \mathbb{D}_\mathrm{KL}\big(\pi_{\theta}(.\vert x) \vert\vert \pi_\mathrm{ref}(. \vert x)\big)
\end{equation}

Denoting the KL-regularized reward as $\Tilde{r}(x,y) = r_{\phi}(x,y)-\beta \log \frac{\pi_{\theta}(y \vert x)}{\pi_\mathrm{ref}(y \vert x)}$, we can express the RL objective compactly as

\begin{equation}
  \max_{\theta} \mathbb{E}_{x \sim \mathcal{D}_x, y \sim \pi_{\theta}(.\vert x)}[\Tilde{r}(x,y)]
\end{equation}

The RL objective can be optimized using a variety of reinforcement learning algorithms, such as Proximal Policy Optimization (PPO) \citep{schulman2017proximalpolicyoptimizationalgorithms}, Group Relative Policy Optimization (GRPO) \citep{shao2024deepseekmathpushinglimitsmathematical}, or REINFORCE \citep{williams1992simple,ahmadian2024basicsrevisitingreinforcestyle}.

\section{Entropic Regularization of Optimal Tranport and the Sinkhorn Algorithm}
\label{supp:ot}

The unregularized optimal transport problem in \Cref{eq:kantorovich-ot} can be solved using linear programming \citep{dantzig2016linear, dvurechensky2018computational}, but this can be computationally expensive when the number of particles is large. Hence, an entropy regularization term is typically added to the objective,
which allows for efficient optimization using the Sinkhorn algorithm \citep{cuturi2013sinkhorn}. Specifically, we define the entropic regularized OT problem between distributions $\mu$ and $\nu$ as:
\begin{equation}
  \mathrm{OT}_{c}^\chi(\mu,\nu) \;=\; \min_{P\in\Pi(\mu,\nu)}\;\langle P,C\rangle - \chi H(P), \quad H(P)\;=\;-\sum_{i,j} P_{ij}\log P_{ij},
  \label{eq:entropic-ot}
\end{equation}
for a regularization parameter $\chi>0$ and a cost matrix $C$. Recall that $\Pi(\mu,\nu)$ represents the set of admissible couplings as stated in~\eqref{eq:transport-plans} (and defined by the marginal of $\mu$ and $\nu$). This problem is strictly convex in  $P$  and has a unique minimizer  $P^\star$. More importantly, it can be solved efficiently using the Sinkhorn algorithm \citep{cuturi2013sinkhorn}, which iteratively updates scaling vectors to enforce the marginal constraints. The resulting transport plan  $P^\star$  can then be used to compute the optimal transport objective and its gradient with respect to the policy parameters  $\theta$. We now decribe how to compute the optimal transport objective.

Let the marginals of $\mu$ and $\nu$ be the vectors $a$ and $b$ respectively. Define the Gibbs kernel
\begin{equation}K_{ij} \;=\; e^{-C_{ij}/\chi}.
  \label{eq:gibbs-kernel}
\end{equation}The optimal transport plan can be expressed in closed form as
\begin{equation}P^\star \;=\; \mathrm{diag}(u) K \mathrm{diag}(v),
  \label{eq:optimal-plan}
\end{equation}where  $u \in \mathbb{R}^N$, $v\in\mathbb{R}^M$  are scaling vectors that can be computed using the Sinkhorn iterations
\begin{equation}u^{(t+1)} \;=\; \frac{a}{Kv^{(t)}}, \qquad v^{(t+1)} \;=\; \frac{b}{K^\top u^{(t+1)}},
  \label{eq:sinkhorn-iterations}
\end{equation}starting from some positive initialization (e.g.  $u^{(0)}=v^{(0)}=\mathbf 1$). The iterations are guaranteed to converge to the unique optimal plan  $P^\star$  that satisfies the marginal constraints \citep{peyre2020computationaloptimaltransport, cuturi2013sinkhorn}. Once we have  $P^\star$ , we can compute the FSD objective as
\begin{equation}
  \mathcal{L}_{\mathrm{FSD}}^\chi(\mu,\nu) \;=\; \langle P^\star,C\rangle - \chi H(P^\star).
  \label{eq:fsd-loss-sinkhorn}
\end{equation}

Because the algorithm consists only of a fixed number of matrix-vector multiplications and divisions, gradients can be propagated through the iterations using automatic differentiation in frameworks like PyTorch \citep{NEURIPS2019_9015} or JAX \citep{jax2018github}, making the Sinkhorn algorithm end to end differentiable.

\section{FSD}
\label{supp:FSD}

\begin{theorem}[from \citet{melnyk2024distributional}]
  \label{thm:fsd_as_ot}
  Let $h : \mathbb{R} \to \mathbb{R}_{+}$ be a convex function, and let
  $X$ and $Y$ be real-valued random variables with probability measures
  $\mu_X$ and $\mu_Y$, respectively. Denote by
  $Q_X, Q_Y : [0,1] \to \mathbb{R}$ their (left-continuous) quantile functions.

  Then,
  \begin{equation}
    \int_0^1 h\big(Q_X(q) - Q_Y(q)\big)\, dq
    =
    \min_{\gamma \in \Pi(\mu_X,\mu_Y)}
    \int_{\mathbb{R}^2} h(x-y)\, d\gamma(x,y),
  \end{equation}
  where $\Pi(\mu_X,\mu_Y)$ denotes the set of all couplings
  (joint probability measures) on $\mathbb{R}^2$ with marginals
  $\mu_X$ and $\mu_Y$. Moreover, if $h$ is strictly convex, the minimizing transport plan $\gamma^\star$ is unique.
\end{theorem}

\begin{corollary}[FSD as asymmetric optimal transport]
  \label{cor:fsd_as_ot}
  Let $X,Y$ be real-valued random variables with measures
  $\mu_X$ and $\mu_Y$, and define the FSD objective
  \[
    \mathcal{L}_{\mathrm{FSD}}(X,Y)
    :=
    \int_0^1 \big(Q_Y(q) - Q_X(q)\big)_+ \, dq.
  \]
  Then
  \[
    \mathcal{L}_{\mathrm{FSD}}(X,Y)
    =
    \min_{\gamma \in \Pi(\mu_X,\mu_Y)}
    \int_{\mathbb{R}^2} (y-x)_+ \, d\gamma(x,y),
  \]
  that is, the FSD objective coincides with the optimal transport cost under the
  asymmetric convex cost function $c(x,y)=(y-x)_+$.
\end{corollary}

\section{Derivation of Theorem~\ref{thm:sard_pg}}\label{app:sard_pg_derivation}

\subsection{RAD Policy Gradient}
\label{sec:sard-pol-grad}


We represent the cost distribution via a particle approximation in quantile space. Let $\{ q_i \}_{i=1}^{N}$ denote an ordered set of cost quantiles satisfying
\[
  q_1 \leq q_2 \leq \dots \leq q_N .
\]
We approximate the induced cost distribution by the empirical measure
\[
  \hat{\mu} = \frac{1}{N} \sum_{i=1}^{N} \delta_{q_i},
\]
where $\delta_{q_i}$ denotes the Dirac measure located at $q_i$. In practice, this approximation is obtained by sampling generations from the policy, evaluating their associated costs, and computing empirical quantiles from the resulting samples. These quantiles serve as the particles defining the discrete representation of the distribution.

Let $\hat{\mu}_{\pi_{\theta}}$ and $\hat{\mu}_{\pi_{\mathrm{ref}}}$ denote the empirical cost distributions of the policy $\pi_{\theta}$ and $\pi_{\mathrm{ref}}$ respectively. Note that we represent $C_{\pi_{\theta}}$ and $C_{\pi_{\mathrm{ref}}}$ using their quantiles $\{q_{i}(\pi_{\theta})\}_{i=1}^{N}$ and $\{q_{i}(\pi_{\mathrm{ref}})\}_{i=1}^{N}$ respectively. The gradient of the FSD objective can be expressed using the chain rule as follows.
\begin{align}
  \nabla_{\theta} \mathcal{L}_{\mathrm{FSD}}\!\big(\hat{\mu}_{\pi_{\theta}},\,\hat{\mu}_{\pi_{\mathrm{ref}}}\big) &= \nabla_{\theta} \mathcal{L}_{\mathrm{FSD}}\big(\{q_{i}(\pi_{\theta})\}_{i=1}^{N}, \{q_{i}(\pi_{\mathrm{ref}})\}_{i=1}^{N}\big)\\
  &= \sum_{i=1}^{N} \frac{\partial \mathcal{L}_{\mathrm{FSD}}}{\partial q_i(\pi_{\theta})} \frac{\partial q_i(\pi_{\theta})}{\partial \theta} = \Big\langle \nabla_{q(\pi_{\theta})}\mathcal{L}_{\mathrm{FSD}}, \nabla_{\theta} q(\pi_{\theta}) \Big\rangle
\end{align}

We denote the gradient vector $\nabla_{q(\pi_{\theta})}\mathcal{L}_{\mathrm{FSD}}$ as the particle gradient. This gradient vector tells us how the FSD objective changes when the particles (quantiles in our setting) used to approximate the empirical distribution of costs of the policy generations are perturbed slightly. The other gradient vector $\nabla_{\theta} q(\pi_{\theta})$ is the quantile gradient, which tells us how the quantiles of the cost distribution change, when the policy parameters are perturbed slightly.

\paragraph{Estimating the Particle Gradient} The FSD objective can be reinterpreted as an optimal transport problem with an asymmetric cost
$c(x,y) = (y - x)_{+}$ as decribed in Corollary~\ref{cor:fsd_as_ot}. The particle gradient can then be viewed as the sensitivity of the optimal transport objective to infinitesimal perturbations of the particle locations.

The classical optimal transport problem (as described in~\eqref{eq:kantorovich-ot}) between $\hat{\mu}_{\pi_\theta}$ and $\hat{\mu}_{\pi_{\mathrm{ref}}}$ is
\begin{equation}
  \mathrm{OT}_c(\hat{\mu}_{\pi_{\theta}},\,\hat{\mu}_{\pi_{\mathrm{ref}}}) := \min_{P \in \Pi(\hat{\mu}_{\pi_{\theta}},\,\hat{\mu}_{\pi_{\mathrm{ref}}})} \langle P, C \rangle
  = \min_{P \in \Pi(\hat{\mu}_{\pi_{\theta}},\,\hat{\mu}_{\pi_{\mathrm{ref}}})} \sum_{i=1}^N \sum_{j=1}^M P_{ij}\, c(x_i, y_j).
  \label{eq:kantorovich-ot-supp}
\end{equation}
where $C_{ij} = c(x_i, y_j)$ is the the cost matrix. Recall that $\Pi(\hat{\mu}_{\pi_{\theta}},\,\hat{\mu}_{\pi_{\mathrm{ref}}}) $ is the set of admissible couplings defined as $\Pi(\hat{\mu}_{\pi_{\theta}},\,\hat{\mu}_{\pi_{\mathrm{ref}}}) := \Bigl\{ P \in \mathbb{R}_+^{N \times N} \;:\; P \mathbf{1}_M = a, \;\; P^\top \mathbf{1}_N = b \Bigr\}.$ The vectors $a$ and $b$ represent the marginals. In our case, the marginals of the empirical cost distributions are $\frac{1}{N}$. 

The classical optimal transport objective as described in~\eqref{eq:kantorovich-ot-supp} is not smooth with respect to particle positions. It is also non-differentiable and small perturbations in particle locations can induce discontinuous changes in the optimal coupling $P^*$. To address these issues, we instead consider the entropy-regularized optimal transport problem as described in~\eqref{eq:entropic-ot}. The regularized objective
\[
  \mathcal{L}_{\mathrm{FSD}}^{\chi}(\hat{\mu}_{\pi_{\theta}},\,\hat{\mu}_{\pi_{\mathrm{ref}}})
  =
  OT^{\chi}_{(y-x)_{+}}\!\left(\hat{\mu}_{\pi_{\theta}},\,\hat{\mu}_{\pi_{\mathrm{ref}}}\right)
\]
adds a strictly convex entropic penalty to the transport plan, yielding a smooth objective that is differentiable with respect to the particle positions. The solution can be computed efficiently using the Sinkhorn algorithm, enabling end-to-end differentiation as described in Section~\ref{supp:ot}.

The regularized objective $\mathcal{L}_{\mathrm{FSD}}^{\chi}$ is a smooth approximation of the original FSD objective $\mathcal{L}_{\mathrm{FSD}}$.
\[
  \mathcal{L}_{\mathrm{FSD}}^{\chi}
  \;\longrightarrow\;
  \mathcal{L}_{\mathrm{FSD}}
  \quad \text{as } \chi \to 0.
\]
In practice, $\chi$ is chosen to balance approximation bias and numerical stability.

\paragraph{Estimating the Quantile Gradient} The quantile gradient informs us how all the quantiles in our empirical distribution of costs of policy generations, change when the policy parameters are perturbed slightly. We consider only the quantile gradient of a single quantile here. The quantile gradient wrt all quantiles can be computed in a similiar manner. Let $F_{\theta}$ be the cumulative distribution function (CDF) of the cost distribution of the model outputs under policy $\pi_{\theta}$. The quantile gradient of a particular quantile $q_i$ is canonically given by $$\nabla_\theta q_i(\pi_\theta)= -\frac{\nabla_\theta F_\theta(t)\vert_{t=q_i}}{f_\theta(q_i)}.$$ Since the PDF in the denominator is unknown, the derivative can be approximated as \citep{li2024tiltedquantilegradientupdates}
\begin{equation}
  \nabla_{\theta} q_{i}(\pi_{\theta}) \approx -\nabla_{\theta} F_{\theta}(q) \Big\vert_{q = q_{i}(\pi_{\theta})}
\end{equation}

The CDF $F_{\theta}$ can be expressed as an expectation and we can use the REINFORCE trick \citep{williams1992simple} to compute the policy gradient of the CDF of the cost distribution of model generations.
\begin{align}
  \nabla_{\theta} F_{\theta}(q) &= \nabla_{\theta} \mathbb{E}_{x \sim \mathcal{D}_{x}, y \sim \pi_{\theta}(. \vert x)} \Big[ \mathds{1}(c_{\psi}(x,y) \leq q)\Big] \\
  &= \mathbb{E}_{x \sim \mathcal{D}_{x}, y \sim \pi_{\theta}(. \vert x)} \Big[ \mathds{1}(c_{\psi}(x,y) \leq q) \nabla_{\theta} \log \pi_{\theta}(y \vert x)\Big]
\end{align}
Recall that the cost distribution is defined via the learned cost model $c_{\psi}(x,y)$ where $x \sim \mathcal{D}_{x}$ is the input prompt and $y \sim \pi_{\theta}(. \vert x)$ is the model output. The quantile gradient for $q_i$ can then be expressed as
\begin{equation}
  \nabla_{\theta} q_{i}(\pi_{\theta}) \approx -\mathbb{E}_{x \sim \mathcal{D}_{x}, y \sim \pi_{\theta}(. \vert x)} \Big[ \mathds{1}(c_{\psi}(x,y) \leq q_i) \nabla_{\theta} \log \pi_{\theta}(y \vert x)\Big]
  \label{eq:quantile-gradient}
\end{equation}

\paragraph{Putting it all together} Using the particle gradient and the quantile gradient, the gradient of the FSD objective with respect to policy parameters can be expressed as

\begin{align}
  \nabla_{\theta} \mathcal{L}_{\mathrm{FSD}}(\hat{\mu}_{\theta}, \hat{\mu}_{\mathrm{ref}}) &= \sum_{i=1}^{N} \frac{\partial \mathcal{L}^{\chi}_{\mathrm{FSD}}}{\partial q_i(\pi_{\theta})} \frac{\partial q_i(\pi_{\theta})}{\partial \theta} \\
  &= \sum_{i=1}^{N} -\frac{\partial \mathcal{L}^{\chi}_{\mathrm{FSD}}}{\partial q_i(\pi_{\theta})} \mathbb{E}_{x \sim \mathcal{D}_{x}, y \sim \pi_{\theta}(. \vert x)} \Big[ \mathds{1}(c_{\psi}(x,y) \leq q_i) \nabla_{\theta} \log \pi_{\theta}(y \vert x)\Big]\\
  &=\mathbb{E}_{x \sim \mathcal{D}_{x}, y \sim \pi_{\theta}(. \vert x)}\Big[ \Big(\sum_{i=1}^{N} -\frac{\partial \mathcal{L}^{\chi}_{\mathrm{FSD}}}{\partial q_i(\pi_{\theta})} \mathds{1}(c_{\psi}(x,y) \leq q_i)\Big) \nabla_{\theta} \log \pi_{\theta}(y \vert x)\Big] \label{eq:fsd-gradient}
\end{align}
Denoting the RAD objective as $\max_{\theta} \min_{\lambda \geq 0} \mathcal{L}(\theta, \lambda)$ the RAD policy gradient can then be expressed as
\begin{equation}
  \nabla_{\theta} \mathcal{L}(\theta, \lambda) = \mathbb{E}_{x \sim \mathcal{D}_{x}, y \sim \pi_{\theta}(. \vert x)}\Big[ \Big( \Tilde{r}(x,y) + \lambda \big(\sum_{i=1}^{N} -\frac{\partial \mathcal{L}^{\chi}_{\mathrm{FSD}}}{\partial q_i(\pi_{\theta})} \mathds{1}(c_{\psi}(x,y) \leq q_i)\big)\Big) \nabla_{\theta} \log \pi_{\theta}(y \vert x)\Big]
  \label{eq:sard-policy-gradient}
\end{equation}

We then use REINFORCE \citep{williams1992simple} with RLOO \citep{Kool2019Buy4R} to optimize the RAD objective using the policy gradient expression in Equation ~\ref{eq:sard-policy-gradient}. For the setting, where we weight quantiles differently in the FSD objective as in Equation ~\ref{eq:weighted-fsd-loss-dfn}, the policy gradient can be expressed as follows, only requiring an additional weighting coefficient $w(q_i)$.

\begin{equation}
  \nabla_{\theta} \mathcal{L}(\theta, \lambda) = \mathbb{E}_{x \sim \mathcal{D}_{x}, y \sim \pi_{\theta}(. \vert x)}\Big[ \Big( \Tilde{r}(x,y) + \lambda \big(\sum_{i=1}^{N} -w(q_i) \frac{\partial \mathcal{L}^{\chi}_{\mathrm{FSD}}}{\partial q_i(\pi_{\theta})} \mathds{1}(c_{\psi}(x,y) \leq q_i)\big)\Big) \nabla_{\theta} \log \pi_{\theta}(y \vert x)\Big]
  \label{eq:weighted-sard-policy-gradient}
\end{equation}


\begin{figure}  
  \centering
  \includegraphics[width=0.9\textwidth]{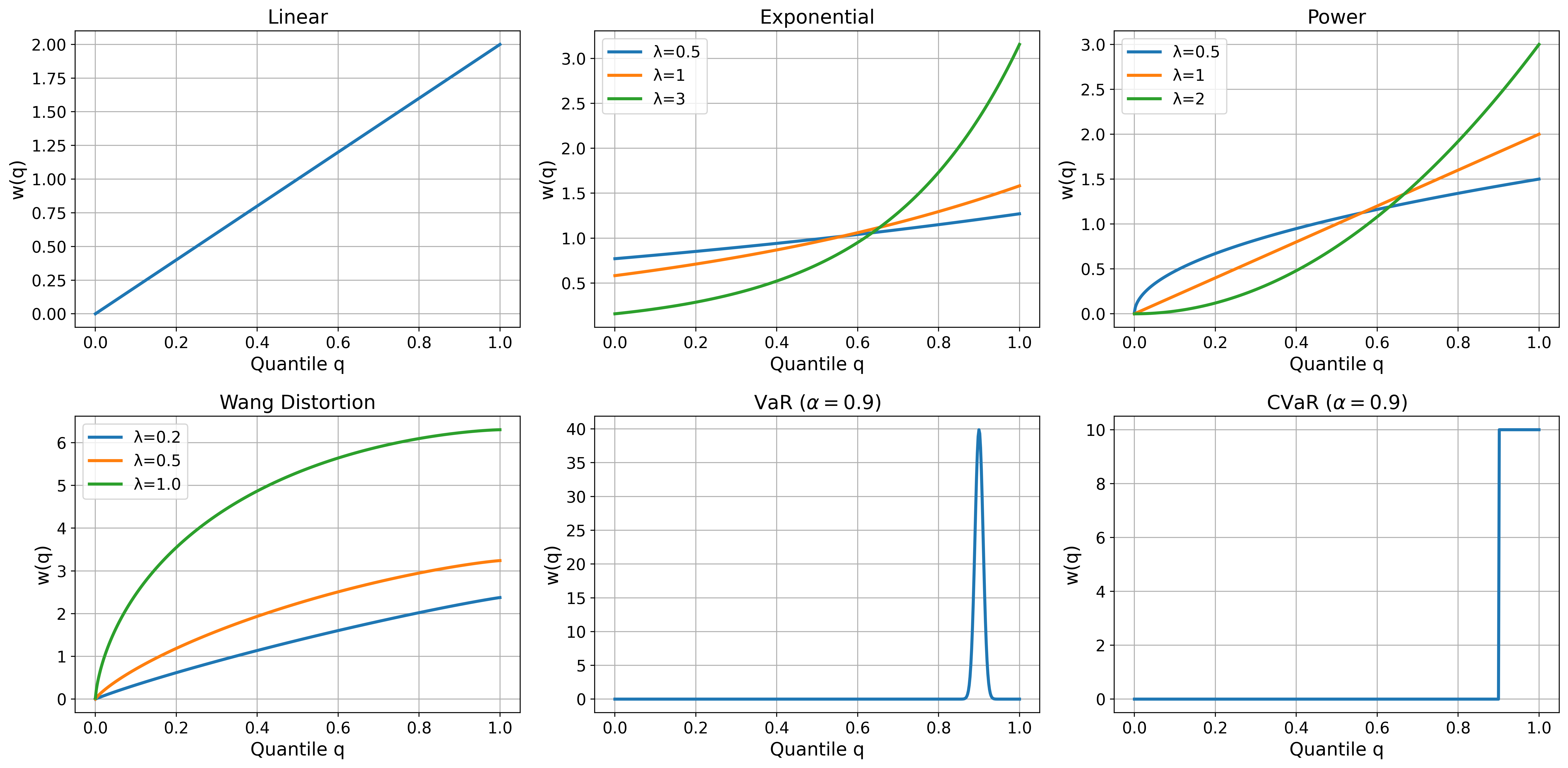}
  \caption{Illustration of spectral weighting functions corresponding to different spectral risk measures (SRMs). For parameterized families, the plots show how the spectral weights vary with the risk-aversion parameter $\lambda$. Note that VaR is represented as a gaussian with very small bandwidth instead of a dirac delta for implementation reasons. \yc{add the params used for wang in the plot} } 
  \label{fig:srm-illustration}
\end{figure}

%% file: main.bib
@article{Christiano2017DeepRL,
  title={Deep Reinforcement Learning from Human Preferences},
  author={Paul Francis Christiano and Jan Leike and Tom B. Brown and Miljan Martic and Shane Legg and Dario Amodei},
  journal={ArXiv},
  year={2017},
  volume={abs/1706.03741},
  url={https://api.semanticscholar.org/CorpusID:4787508}
}

@article{Ouyang2022TrainingLM,
  title={Training language models to follow instructions with human feedback},
  author={Long Ouyang and others},
  journal={ArXiv},
  year={2022},
  volume={abs/2203.02155},
  url={https://api.semanticscholar.org/CorpusID:246426909}
}

@article{Bai2022ConstitutionalAH,
  title={Constitutional {AI}: Harmlessness from {AI} Feedback},
  author={Yuntao Bai and others},
  journal={ArXiv},
  year={2022},
  volume={abs/2212.08073},
  url={https://api.semanticscholar.org/CorpusID:254823489}
}

@article{Jaques2019WayOB,
  title={Way Off-Policy Batch Deep Reinforcement Learning of Implicit Human Preferences in Dialog},
  author={Natasha Jaques and Asma Ghandeharioun and Judy Hanwen Shen and Craig Ferguson and {\`A}gata Lapedriza and Noah J. Jones and Shixiang Shane Gu and Rosalind W. Picard},
  journal={ArXiv},
  year={2019},
  volume={abs/1907.00456},
  url={https://api.semanticscholar.org/CorpusID:195766797}
}

@misc{stiennon2022learningsummarizehumanfeedback,
      title={Learning to summarize from human feedback}, 
      author={Nisan Stiennon and Long Ouyang and Jeff Wu and Daniel M. Ziegler and Ryan Lowe and Chelsea Voss and Alec Radford and Dario Amodei and Paul Christiano},
      year={2022},
      eprint={2009.01325},
      archivePrefix={arXiv},
      primaryClass={cs.CL},
      url={https://arxiv.org/abs/2009.01325}, 
}

@misc{schulman2017proximalpolicyoptimizationalgorithms,
      title={Proximal Policy Optimization Algorithms}, 
      author={John Schulman and Filip Wolski and Prafulla Dhariwal and Alec Radford and Oleg Klimov},
      year={2017},
      eprint={1707.06347},
      archivePrefix={arXiv},
      primaryClass={cs.LG},
      url={https://arxiv.org/abs/1707.06347}, 
}

@inproceedings{Gao2022ScalingLF,
  title={Scaling Laws for Reward Model Overoptimization},
  author={Leo Gao and John Schulman and Jacob Hilton},
  booktitle={International Conference on Machine Learning},
  year={2022},
  url={https://api.semanticscholar.org/CorpusID:252992904}
}

@article{Rafailov2024ScalingLF,
  title={Scaling Laws for Reward Model Overoptimization in Direct Alignment Algorithms},
  author={Rafael Rafailov and Yaswanth Chittepu and Ryan Park and Harshit S. Sikchi and Joey Hejna and Bradley Knox and Chelsea Finn and Scott Niekum},
  journal={ArXiv},
  year={2024},
  volume={abs/2406.02900},
  url={https://api.semanticscholar.org/CorpusID:270257855}
}

@article{gallier2019fundamentals,
  title={Fundamentals of optimization theory with applications to machine learning},
  author={Gallier, Jean and Quaintance, Jocelyn},
  journal={University of Pennsylvania Philadelphia, PA},
  volume={19104},
  year={2019}
}

@misc{qwen2025qwen25technicalreport,
      title={Qwen2.5 Technical Report}, 
      author={Qwen and others},
      year={2025},
      eprint={2412.15115},
      archivePrefix={arXiv},
      primaryClass={cs.CL},
      url={https://arxiv.org/abs/2412.15115}, 
}

@misc{ji2023beavertailsimprovedsafetyalignment,
      title={BeaverTails: Towards Improved Safety Alignment of {LLM} via a Human-Preference Dataset}, 
      author={Jiaming Ji and Mickel Liu and Juntao Dai and Xuehai Pan and Chi Zhang and Ce Bian and Chi Zhang and Ruiyang Sun and Yizhou Wang and Yaodong Yang},
      year={2023},
      eprint={2307.04657},
      archivePrefix={arXiv},
      primaryClass={cs.CL},
      url={https://arxiv.org/abs/2307.04657}, 
}

@misc{zheng2023judgingllmasajudgemtbenchchatbot,
      title={Judging {LLM}-as-a-{J}udge with {MT}-{B}ench and {C}hatbot {A}rena}, 
      author={Lianmin Zheng and Wei-Lin Chiang and Ying Sheng and Siyuan Zhuang and Zhanghao Wu and Yonghao Zhuang and Zi Lin and Zhuohan Li and Dacheng Li and Eric P. Xing and Hao Zhang and Joseph E. Gonzalez and Ion Stoica},
      year={2023},
      eprint={2306.05685},
      archivePrefix={arXiv},
      primaryClass={cs.CL},
      url={https://arxiv.org/abs/2306.05685}, 
}

@book{altman2021constrained,
  title={Constrained Markov decision processes},
  author={Altman, Eitan},
  year={2021},
  publisher={Routledge}
}

@misc{bai2022traininghelpfulharmlessassistant,
      title={Training a Helpful and Harmless Assistant with Reinforcement Learning from Human Feedback}, 
      author={Yuntao Bai and others},
      year={2022},
      eprint={2204.05862},
      archivePrefix={arXiv},
      primaryClass={cs.CL},
      url={https://arxiv.org/abs/2204.05862}
}

@article{dai2023safe,
  title={Safe {RLHF}: Safe reinforcement learning from human feedback},
  author={Dai, Josef and Pan, Xuehai and Sun, Ruiyang and Ji, Jiaming and Xu, Xinbo and Liu, Mickel and Wang, Yizhou and Yang, Yaodong},
  journal={arXiv preprint arXiv:2310.12773},
  year={2023}
}

@article{bradley1952rank,
  title={Rank analysis of incomplete block designs: I. {T}he method of paired comparisons},
  author={Bradley, Ralph Allan and Terry, Milton E},
  journal={Biometrika},
  volume={39},
  number={3/4},
  pages={324--345},
  year={1952},
  publisher={JSTOR}
}

@misc{ahmadian2024basicsrevisitingreinforcestyle,
      title={Back to Basics: Revisiting REINFORCE Style Optimization for Learning from Human Feedback in {LLMs}}, 
      author={Arash Ahmadian and Chris Cremer and Matthias Gallé and Marzieh Fadaee and Julia Kreutzer and Olivier Pietquin and Ahmet Üstün and Sara Hooker},
      year={2024},
      eprint={2402.14740},
      archivePrefix={arXiv},
      primaryClass={cs.LG},
      url={https://arxiv.org/abs/2402.14740}, 
}

@inproceedings{Kool2019Buy4R,
  title={Buy 4 REINFORCE Samples, Get a Baseline for Free!},
  author={Wouter Kool and Herke van Hoof and Max Welling},
  booktitle={DeepRLStructPred@ICLR},
  year={2019},
  url={https://api.semanticscholar.org/CorpusID:198489118}
}

@article{williams1992simple,
  title={Simple statistical gradient-following algorithms for connectionist reinforcement learning},
  author={Williams, Ronald J},
  journal={Machine learning},
  volume={8},
  pages={229--256},
  year={1992},
  publisher={Springer}
}

@misc{shao2024deepseekmathpushinglimitsmathematical,
      title={DeepSeekMath: Pushing the Limits of Mathematical Reasoning in Open Language Models}, 
      author={Zhihong Shao and Peiyi Wang and Qihao Zhu and Runxin Xu and Junxiao Song and Xiao Bi and Haowei Zhang and Mingchuan Zhang and Y. K. Li and Y. Wu and Daya Guo},
      year={2024},
      eprint={2402.03300},
      archivePrefix={arXiv},
      primaryClass={cs.CL},
      url={https://arxiv.org/abs/2402.03300}, 
}

@inproceedings{chittepu2025reinforcement,
  title={Reinforcement learning from human feedback with high-confidence safety guarantees},
  author={Chittepu, Yaswanth and Metevier, Blossom and Schwarzer, Will and Hoag, Austin and Niekum, Scott and Thomas, Philip S},
  booktitle={Reinforcement Learning Conference},
  year={2025}
}

@article{thomas2019preventing,
  title={Preventing undesirable behavior of intelligent machines},
  author={Thomas, Philip S and Castro da Silva, Bruno and Barto, Andrew G and Giguere, Stephen and Brun, Yuriy and Brunskill, Emma},
  journal={Science},
  volume={366},
  number={6468},
  pages={999--1004},
  year={2019},
  publisher={American Association for the Advancement of Science}
}

@misc{peyre2020computationaloptimaltransport,
      title={Computational Optimal Transport}, 
      author={Gabriel Peyré and Marco Cuturi},
      year={2020},
      eprint={1803.00567},
      archivePrefix={arXiv},
      primaryClass={stat.ML},
      url={https://arxiv.org/abs/1803.00567}, 
}

@article{dantzig2016linear,
  title={Linear programming and extensions},
  author={Dantzig, George B},
  year={2016},
  publisher={Princeton university press}
}

@inproceedings{dvurechensky2018computational,
  title={Computational optimal transport: Complexity by accelerated gradient descent is better than by Sinkhorn’s algorithm},
  author={Dvurechensky, Pavel and Gasnikov, Alexander and Kroshnin, Alexey},
  booktitle={International conference on machine learning},
  pages={1367--1376},
  year={2018},
  organization={PMLR}
}

@article{cuturi2013sinkhorn,
  title={Sinkhorn distances: Lightspeed computation of optimal transport},
  author={Cuturi, Marco},
  journal={Advances in neural information processing systems},
  volume={26},
  year={2013}
}

@software{jax2018github,
  author = {James Bradbury and Roy Frostig and Peter Hawkins and Matthew James Johnson and Chris Leary and Dougal Maclaurin and George Necula and Adam Paszke and Jake Vander{P}las and Skye Wanderman-{M}ilne and Qiao Zhang},
  title = {{JAX}: composable transformations of {P}ython+{N}um{P}y programs},
  url = {http://github.com/jax-ml/jax},
  version = {0.3.13},
  year = {2018},
}

@incollection{NEURIPS2019_9015,
title = {PyTorch: An Imperative Style, High-Performance Deep Learning Library},
author = {Paszke, Adam and Gross, Sam and Massa, Francisco and Lerer, Adam and Bradbury, James and Chanan, Gregory and Killeen, Trevor and Lin, Zeming and Gimelshein, Natalia and Antiga, Luca and Desmaison, Alban and Kopf, Andreas and Yang, Edward and DeVito, Zachary and Raison, Martin and Tejani, Alykhan and Chilamkurthy, Sasank and Steiner, Benoit and Fang, Lu and Bai, Junjie and Chintala, Soumith},
booktitle = {Advances in Neural Information Processing Systems 32},
pages = {8024--8035},
year = {2019},
publisher = {Curran Associates, Inc.},
url = {http://papers.neurips.cc/paper/9015-pytorch-an-imperative-style-high-performance-deep-learning-library.pdf}
}

@inproceedings{Santambrogio2015OptimalTF,
  title={Optimal Transport for Applied Mathematicians: Calculus of Variations, PDEs, and Modeling},
  author={Filippo Santambrogio},
  year={2015},
  url={https://api.semanticscholar.org/CorpusID:124181096}
}

@article{melnyk2024distributional,
  title={Distributional preference alignment of llms via optimal transport},
  author={Melnyk, Igor and Mroueh, Youssef and Belgodere, Brian and Rigotti, Mattia and Nitsure, Apoorva and Yurochkin, Mikhail and Greenewald, Kristjan and Navratil, Jiri and Ross, Jarret},
  journal={Advances in Neural Information Processing Systems},
  volume={37},
  pages={104412--104442},
  year={2024}
}

@article{artzner1999coherent,
  title={Coherent measures of risk},
  author={Artzner, Philippe and Delbaen, Freddy and Eber, Jean-Marc and Heath, David},
  journal={Mathematical finance},
  volume={9},
  number={3},
  pages={203--228},
  year={1999},
  publisher={Wiley Online Library}
}

@article{acerbi2002spectral,
  title={Spectral measures of risk: A coherent representation of subjective risk aversion},
  author={Acerbi, Carlo},
  journal={Journal of banking \& finance},
  volume={26},
  number={7},
  pages={1505--1518},
  year={2002},
  publisher={Elsevier}
}

@misc{li2024tiltedquantilegradientupdates,
      title={Tilted Quantile Gradient Updates for Quantile-Constrained Reinforcement Learning}, 
      author={Chenglin Li and Guangchun Ruan and Hua Geng},
      year={2024},
      eprint={2412.13184},
      archivePrefix={arXiv},
      primaryClass={cs.LG},
      url={https://arxiv.org/abs/2412.13184}, 
}

@article{yang2022large,
  title={A large language model for electronic health records},
  author={Yang, Xi and Chen, Aokun and PourNejatian, Nima and Shin, Hoo Chang and Smith, Kaleb E and Parisien, Christopher and Compas, Colin and Martin, Cheryl and Costa, Anthony B and Flores, Mona G and others},
  journal={NPJ digital medicine},
  volume={5},
  number={1},
  pages={194},
  year={2022},
  publisher={Nature Publishing Group UK London}
}

@article{moor2023foundation,
  title={Foundation models for generalist medical artificial intelligence},
  author={Moor, Michael and Banerjee, Oishi and Abad, Zahra Shakeri Hossein and Krumholz, Harlan M and Leskovec, Jure and Topol, Eric J and Rajpurkar, Pranav},
  journal={Nature},
  volume={616},
  number={7956},
  pages={259--265},
  year={2023},
  publisher={Nature Publishing Group UK London}
}

@article{katz2024gpt,
  title={Gpt-4 passes the bar exam},
  author={Katz, Daniel Martin and Bommarito, Michael James and Gao, Shang and Arredondo, Pablo},
  journal={Philosophical Transactions of the Royal Society A: Mathematical, Physical and Engineering Sciences},
  volume={382},
  number={2270},
  year={2024},
  publisher={The Royal Society}
}

@article{kasneci2023chatgpt,
  title={ChatGPT for good? On opportunities and challenges of large language models for education},
  author={Kasneci, Enkelejda and Se{\ss}ler, Kathrin and K{\"u}chemann, Stefan and Bannert, Maria and Dementieva, Daryna and Fischer, Frank and Gasser, Urs and Groh, Georg and G{\"u}nnemann, Stephan and H{\"u}llermeier, Eyke and others},
  journal={Learning and individual differences},
  volume={103},
  pages={102274},
  year={2023},
  publisher={Elsevier}
}

@article{kung2023performance,
  title={Performance of ChatGPT on USMLE: potential for AI-assisted medical education using large language models},
  author={Kung, Tiffany H and Cheatham, Morgan and Medenilla, Arielle and Sillos, Czarina and De Leon, Lorie and Elepa{\~n}o, Camille and Madriaga, Maria and Aggabao, Rimel and Diaz-Candido, Giezel and Maningo, James and others},
  journal={PLoS digital health},
  volume={2},
  number={2},
  pages={e0000198},
  year={2023},
  publisher={Public Library of Science}
}

@inproceedings{gehman2020realtoxicityprompts,
  title={Realtoxicityprompts: Evaluating neural toxic degeneration in language models},
  author={Gehman, Samuel and Gururangan, Suchin and Sap, Maarten and Choi, Yejin and Smith, Noah A},
  booktitle={Findings of the association for computational linguistics: EMNLP 2020},
  pages={3356--3369},
  year={2020}
}

@article{weidinger2021ethical,
  title={Ethical and social risks of harm from language models},
  author={Weidinger, Laura and Mellor, John and Rauh, Maribeth and Griffin, Conor and Uesato, Jonathan and Huang, Po-Sen and Cheng, Myra and Glaese, Mia and Balle, Borja and Kasirzadeh, Atoosa and others},
  journal={arXiv preprint arXiv:2112.04359},
  year={2021}
}

@article{ganguli2022red,
  title={Red teaming language models to reduce harms: Methods, scaling behaviors, and lessons learned},
  author={Ganguli, Deep and Lovitt, Liane and Kernion, Jackson and Askell, Amanda and Bai, Yuntao and Kadavath, Saurav and Mann, Ben and Perez, Ethan and Schiefer, Nicholas and Ndousse, Kamal and others},
  journal={arXiv preprint arXiv:2209.07858},
  year={2022}
}

@article{glaese2022improving,
  title={Improving alignment of dialogue agents via targeted human judgements},
  author={Glaese, Amelia and McAleese, Nat and Tr{\k{e}}bacz, Maja and Aslanides, John and Firoiu, Vlad and Ewalds, Timo and Rauh, Maribeth and Weidinger, Laura and Chadwick, Martin and Thacker, Phoebe and others},
  journal={arXiv preprint arXiv:2209.14375},
  year={2022}
}

@article{bai2022constitutional,
  title={Constitutional ai: Harmlessness from ai feedback},
  author={Bai, Yuntao and Kadavath, Saurav and Kundu, Sandipan and Askell, Amanda and Kernion, Jackson and Jones, Andy and Chen, Anna and Goldie, Anna and Mirhoseini, Azalia and McKinnon, Cameron and others},
  journal={arXiv preprint arXiv:2212.08073},
  year={2022}
}

@article{ouyang2022training,
  title={Training language models to follow instructions with human feedback},
  author={Ouyang, Long and Wu, Jeffrey and Jiang, Xu and Almeida, Diogo and Wainwright, Carroll and Mishkin, Pamela and Zhang, Chong and Agarwal, Sandhini and Slama, Katarina and Ray, Alex and others},
  journal={Advances in neural information processing systems},
  volume={35},
  pages={27730--27744},
  year={2022}
}

@article{touvron2023llama,
  title={Llama 2: Open foundation and fine-tuned chat models},
  author={Touvron, Hugo and Martin, Louis and Stone, Kevin and Albert, Peter and Almahairi, Amjad and Babaei, Yasmine and Bashlykov, Nikolay and Batra, Soumya and Bhargava, Prajjwal and Bhosale, Shruti and others},
  journal={arXiv preprint arXiv:2307.09288},
  year={2023}
}

@article{mu2024rule,
  title={Rule based rewards for language model safety},
  author={Mu, Tong and Helyar, Alec and Heidecke, Johannes and Achiam, Joshua and Vallone, Andrea and Kivlichan, Ian and Lin, Molly and Beutel, Alex and Schulman, John and Weng, Lilian},
  journal={Advances in Neural Information Processing Systems},
  volume={37},
  pages={108877--108901},
  year={2024}
}

@inproceedings{dai2023learning,
  title={Learning to optimize with stochastic dominance constraints},
  author={Dai, Hanjun and Xue, Yuan and He, Niao and Wang, Yixin and Li, Na and Schuurmans, Dale and Dai, Bo},
  booktitle={International Conference on Artificial Intelligence and Statistics},
  pages={8991--9009},
  year={2023},
  organization={PMLR}
}

@article{cen2024beyond,
  title={Beyond Expectations: Learning with Stochastic Dominance Made Practical},
  author={Cen, Shicong and Mei, Jincheng and Dai, Hanjun and Schuurmans, Dale and Chi, Yuejie and Dai, Bo},
  journal={arXiv preprint arXiv:2402.02698},
  year={2024}
}

@article{dentcheva2003optimization,
  title={Optimization with stochastic dominance constraints},
  author={Dentcheva, Darinka and Ruszczynski, Andrzej},
  journal={SIAM Journal on Optimization},
  volume={14},
  number={2},
  pages={548--566},
  year={2003},
  publisher={SIAM}
}

@inproceedings{farajzadeh2025imitation,
  title={Imitation beyond expectation using pluralistic stochastic dominance},
  author={Farajzadeh, Ali and Saeed, Danyal and Abbas, Syed M and Shah, Rushit N and Saha, Aadirupa and Ziebart, Brian D},
  booktitle={The Thirty-ninth Annual Conference on Neural Information Processing Systems},
  year={2025}
}

@article{xu2020recipes,
  title={Recipes for safety in open-domain chatbots},
  author={Xu, Jing and Ju, Da and Li, Margaret and Boureau, Y-Lan and Weston, Jason and Dinan, Emily},
  journal={arXiv preprint arXiv:2010.07079},
  year={2020}
}

@article{thoppilan2022lamda,
  title={Lamda: Language models for dialog applications},
  author={Thoppilan, Romal and De Freitas, Daniel and Hall, Jamie and Shazeer, Noam and Kulshreshtha, Apoorv and Cheng, Heng-Tze and Jin, Alicia and Bos, Taylor and Baker, Leslie and Du, Yu and others},
  journal={arXiv preprint arXiv:2201.08239},
  year={2022}
}

@article{ziegler2022adversarial,
  title={Adversarial training for high-stakes reliability},
  author={Ziegler, Daniel M and Nix, Seraphina and Chan, Lawrence and Bauman, Tim and Schmidt-Nielsen, Peter and Lin, Tao and Scherlis, Adam and Nabeshima, Noa and Weinstein-Raun, Ben and de Haas, Daniel and others},
  journal={arXiv preprint arXiv:2205.01663},
  year={2022}
}

@misc{mazeika2024harmbenchstandardizedevaluationframework,
      title={HarmBench: A Standardized Evaluation Framework for Automated Red Teaming and Robust Refusal}, 
      author={Mantas Mazeika and Long Phan and Xuwang Yin and Andy Zou and Zifan Wang and Norman Mu and Elham Sakhaee and Nathaniel Li and Steven Basart and Bo Li and David Forsyth and Dan Hendrycks},
      year={2024},
      eprint={2402.04249},
      archivePrefix={arXiv},
      primaryClass={cs.LG},
      url={https://arxiv.org/abs/2402.04249}, 
}

@misc{pearce2021asleepkeyboardassessingsecurity,
      title={Asleep at the Keyboard? Assessing the Security of GitHub Copilot's Code Contributions}, 
      author={Hammond Pearce and Baleegh Ahmad and Benjamin Tan and Brendan Dolan-Gavitt and Ramesh Karri},
      year={2021},
      eprint={2108.09293},
      archivePrefix={arXiv},
      primaryClass={cs.CR},
      url={https://arxiv.org/abs/2108.09293}, 
}

@misc{perez2022redteaminglanguagemodels,
      title={Red Teaming Language Models with Language Models}, 
      author={Ethan Perez and Saffron Huang and Francis Song and Trevor Cai and Roman Ring and John Aslanides and Amelia Glaese and Nat McAleese and Geoffrey Irving},
      year={2022},
      eprint={2202.03286},
      archivePrefix={arXiv},
      primaryClass={cs.CL},
      url={https://arxiv.org/abs/2202.03286}, 
}

@misc{alpaca,
  author = {Rohan Taori and Ishaan Gulrajani and Tianyi Zhang and Yann Dubois and Xuechen Li and Carlos Guestrin and Percy Liang and Tatsunori B. Hashimoto },
  title = {Stanford {A}lpaca: An Instruction-following {LLaMA} model},
  year = {2023},
  publisher = {GitHub},
  journal = {GitHub repository},
  howpublished = {\url{https://github.com/tatsu-lab/stanford_alpaca}},
}

@article{wang1996premium,
  title={Premium calculation by transforming the layer premium density},
  author={Wang, Shaun},
  journal={ASTIN Bulletin: The Journal of the IAA},
  volume={26},
  number={1},
  pages={71--92},
  year={1996},
  publisher={Cambridge University Press}
}

@misc{openai2024gpt4omini,
  title={GPT-4o mini: Advancing cost-efficient intelligence},
  author={OpenAI},
  year={2024},
  howpublished={\url{https://openai.com/index/gpt-4o-mini-advancing-cost-efficient-intelligence/}},
}

@article{dowd2008spectral,
  title={Spectral risk measures: properties and limitations},
  author={Dowd, Kevin and Cotter, John and Sorwar, Ghulam},
  journal={Journal of Financial Services Research},
  volume={34},
  number={1},
  pages={61--75},
  year={2008},
  publisher={Springer}
}

@article{adam2008spectral,
  title={Spectral risk measures and portfolio selection},
  author={Adam, Alexandre and Houkari, Mohamed and Laurent, Jean-Paul},
  journal={Journal of Banking \& Finance},
  volume={32},
  number={9},
  pages={1870--1882},
  year={2008},
  publisher={Elsevier}
}

@article{flamary2021pot,
  author  = {R{\'e}mi Flamary and Nicolas Courty and Alexandre Gramfort and Mokhtar Z. Alaya and Aur{\'e}lie Boisbunon and Stanislas Chambon and Laetitia Chapel and Adrien Corenflos and Kilian Fatras and Nemo Fournier and L{\'e}o Gautheron and Nathalie T.H. Gayraud and Hicham Janati and Alain Rakotomamonjy and Ievgen Redko and Antoine Rolet and Antony Schutz and Vivien Seguy and Danica J. Sutherland and Romain Tavenard and Alexander Tong and Titouan Vayer},
  title   = {POT: Python Optimal Transport},
  journal = {Journal of Machine Learning Research},
  year    = {2021},
  volume  = {22},
  number  = {78},
  pages   = {1-8},
  url     = {http://jmlr.org/papers/v22/20-451.html}
}
